\documentclass[lettersize,journal]{IEEEtran}

\usepackage{amsmath,amsfonts}
\usepackage{algorithmic}
\usepackage{algorithm}
\usepackage{array}
\usepackage[caption=false,font=normalsize,labelfont=sf,textfont=sf]{subfig}
\usepackage{textcomp}
\usepackage{stfloats}
\usepackage{url}
\usepackage{array}
\usepackage{makecell}
\usepackage{multirow}
\usepackage{verbatim}
\usepackage{graphicx}
\usepackage{cite}
\usepackage{hyperref}
\usepackage{bm}
\usepackage{mathtools}
\usepackage{enumitem}
\hypersetup{colorlinks=true,
            linkcolor=green,
            urlcolor=black,
            citecolor=green}
\newcommand{\PreserveBackslash}[1]{\let\temp=\\#1\let\\=\temp}
\newcolumntype{C}[1]{>{\PreserveBackslash\centering}p{#1}}
\newcolumntype{R}[1]{>{\PreserveBackslash\raggedleft}p{#1}}
\newcolumntype{L}[1]{>{\PreserveBackslash\raggedright}p{#1}}

\newcommand{\zsyrevise}[1]{\textcolor{black}{#1}}

\begin{document}

\title{DifFace: Blind Face Restoration with Diffused Error Contraction}

\author{Zongsheng Yue,~Chen Change Loy,~\IEEEmembership{Senior Member,~IEEE}
\thanks{Z. Yue and C. C. Loy are with S-Lab, Nanyang Technological University (NTU), Singapore (E-mail: zsyue@gmail.com, ccloy@ntu.edu.sg).}
\thanks{C. C. Loy is the corresponding author.}
}

\markboth{}%
{Shell \MakeLowercase{\textit{et al.}}: A Sample Article Using IEEEtran.cls for IEEE Journals}


\maketitle

\begin{abstract}
While deep learning-based methods for blind face restoration have achieved unprecedented success, they still suffer from two major limitations. First, most of them deteriorate when facing complex degradations out of their training data. Second, these methods require multiple constraints, e.g., fidelity, perceptual, and adversarial losses, which require laborious hyper-parameter tuning to stabilize and balance their influences. In this work, we propose a novel method named \textit{DifFace} that is capable of coping with unseen and complex degradations more gracefully without complicated loss designs. The key of our method is to establish a posterior distribution from the observed low-quality (LQ) image to its high-quality (HQ) counterpart. In particular, we design a transition distribution from the LQ image to the intermediate state of a pre-trained diffusion model and then gradually transmit from this intermediate state to the HQ target by recursively applying a pre-trained diffusion model. The transition distribution only relies on a restoration backbone that is trained with $L_1$ loss on some synthetic data, which favorably avoids the cumbersome training process in existing methods. Moreover, the transition distribution can contract the error of the restoration backbone and thus makes our method more robust to unknown degradations. Comprehensive experiments show that \textit{DifFace} is superior to current state-of-the-art methods, especially in cases with severe degradations. Code and model are available at \url{https://github.com/zsyOAOA/DifFace}.
\end{abstract}


\section{Introduction}
\IEEEPARstart{B}lind face restoration (BFR) aims at recovering a high-quality (HQ) image from its low-quality (LQ) counterpart, which usually suffers from complex degradations, such as noise, blurring, and downsampling. BFR is an extremely ill-posed inverse problem as multiple HQ solutions may exist for any given LQ image.

Approaches for BFR have been dominated by deep learning-based methods~\cite{wang2021towards,tu2021joint,feihong2022toward,gu2022vqfr}. The main idea of them is to learn a mapping, usually parameterized as a deep neural network, from the LQ images to the HQ ones based on a large amount of pre-collected LQ/HQ image pairs. In most cases, these image pairs are synthesized by assuming a degradation model that often deviates from the real one. Most existing methods are sensitive to such a deviation and thus suffer a dramatic performance drop when encountering mismatched degradations in real scenarios.
\begin{figure}[t]
    \centering
    \includegraphics[width=\columnwidth]{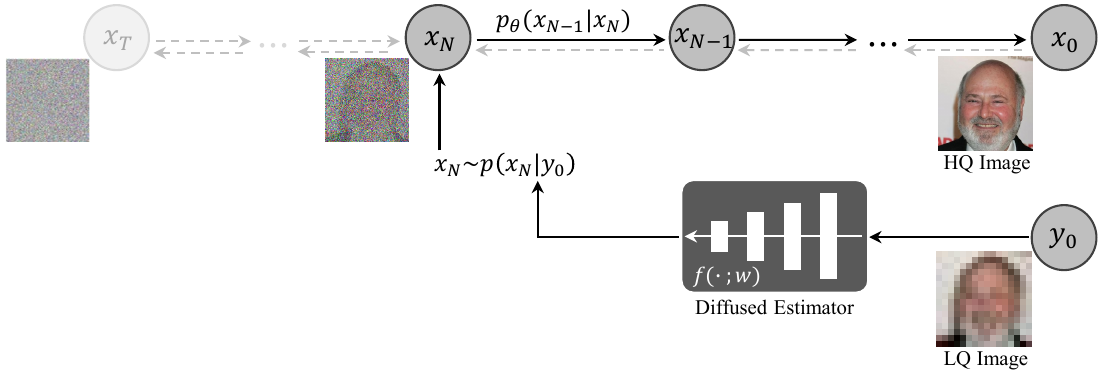}
    \vspace{-6mm}
    \caption{Overview of the proposed method. The solid lines denote the whole inference pipeline of our method. For ease of comparison, we also mark out the forward and reverse processes of the diffusion model by dotted lines.}
    \label{fig:framework}
\end{figure}

Various constraints or priors have been designed to mitigate the influence of such a deviation and improve the restoration quality. The $L_2$ (or $L_1$) loss is commonly used to ensure fidelity, although these pixel-wise losses are known to favor the prediction of an average (or a median) over plausible solutions.
Recent BFR methods also introduce the adversarial loss~\cite{goodfellow2014generative} and the perceptual loss~\cite{johnson2016perceptual,zhang2018unreasonable} to achieve more realistic results.
Besides, some existing methods also exploit face-specific priors to further constrain the restored solution, e.g., face landmarks~\cite{chen2018fsrnet}, facial components~\cite{li2020blind}, and generative priors~\cite{chan2021glean,pan2021exploiting,wang2021towards,yang2021gan,toward2022zhou}.
Considering so many constraints together makes the training unnecessarily complicated, often requiring laborious hyper-parameter tuning to make a trade-off among these constraints.
Worse, the notorious instability of adversarial loss makes the training more challenging. 

\begin{figure*}[t]
    \centering
    \includegraphics[width=\linewidth]{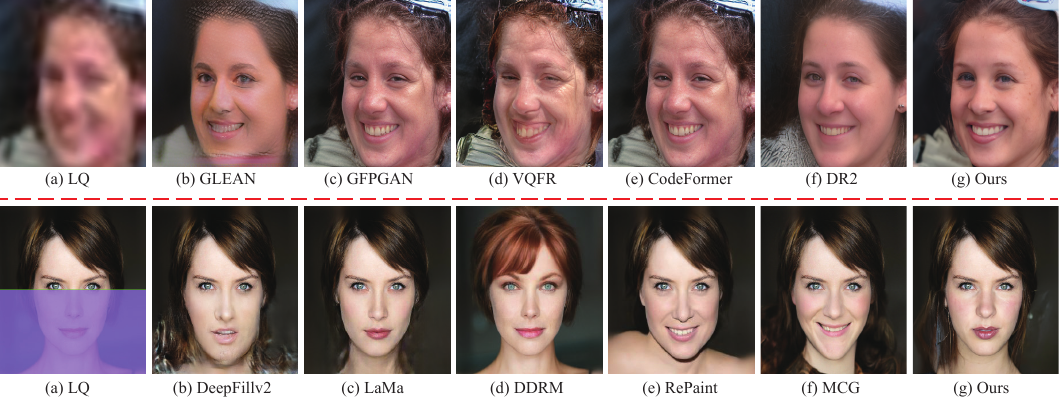}
    \vspace{-6mm}
    \caption{\zsyrevise{Comparative results of the proposed method to recent state-of-the-art approaches on the task of blind face restoration (top row) and face image inpainting (bottom row). Note that the masked area is highlighted using a purple color in the example of face inpainting.}}
    \label{fig:method_visulization}
\end{figure*}

In this work, we devise a novel BFR method \textit{DifFace}, inspired by the recent success of diffusion model in image generation~\cite{sohl2015deep,dhariwal2021diffusion}.
Our method does not require multiple constraints in training, and yet is capable of dealing with unknown and complex degradations. 
Importantly, we leverage the rich image priors and strong generative capability of a pretrained diffusion model without retraining it on any manually assumed degradations. 

To this end, we establish a posterior distribution $p(\bm{x}_0|\bm{y}_0)$, aiming to infer the HQ image $\bm{x}_0$ conditioned on its LQ counterpart $\bm{y}_0$. Due to the complex degradations, solving for this posterior is non-trivial in blind restoration.
Our solution to this problem, as depicted in Fig.~\ref{fig:framework}, is to approximate this posterior distribution by a transition distribution $p(\bm{x}_N|\bm{y}_0)$, where $\bm{x}_N$ is a diffused version of the desirable HQ image $\bm{x}_0$, followed with a reverse Markov chain that estimates $\bm{x}_0$ from $\bm{x}_N$. The transition distribution $p(\bm{x}_N|\bm{y}_0)$ can be built up by introducing a neural network that is trained just using $L_1$ loss, and the reverse Markov chain is readily borrowed from a pretrained diffusion model containing abundant image priors.

The constructed transition distribution is appealing in that it is well motivated by an important observation in DDPM~\cite{ho2020denoising}, where data is destroyed by re-scaling it with a factor of less than 1 and adding noise in the diffusion process. Bringing this notion to our context, the residual between $\bm{x}_0$ and $\bm{y}_0$ is also contracted by this factor after diffusion.
Our framework uniquely leverages this property by inferring the intermediate diffused variable $\bm{x}_N$ (where $N < T$) from the LQ image $\bm{y}_0$, of which the residual to HQ image $\bm{x}_0$ is reduced. And then from this intermediate state we infer the desirable $\bm{x}_0$. There are several advantages of doing so: i) our solution is more efficient than the full reverse diffusion process from $\bm{x}_T$ to $\bm{x}_0$, ii) we do not need to retrain the diffusion model from scratch, and iii) we can still take advantage of the pre-trained diffusion model via the reverse Markov chain from $\bm{x}_N$ to $\bm{x}_0$.

SR3~\cite{saharia2022image} also exploits the potentials of the diffusion model for blind restoration. It feeds the LQ image into the diffusion model as a condition to guide the restoration at each timestep. This requires one to retrain the diffusion model from scratch on pre-collected training data. Hence, it would still suffer from the issue of degradation mismatch when dealing with real-world data. Unlike SR3, our method does not need to train the diffusion model from scratch but sufficiently leverages the prior knowledge of the pretrained diffusion model. The unique design on transition distribution $p(\bm{x}_N|\bm{y}_0)$ further allows us to better cope with unknown degradations.

In summary, the contributions of this work are as follows:
\begin{itemize}
    \item We devise a new diffusion-based BFR approach to cope with severe and unknown degradations. Formulating the posterior distribution as a Markov chain that starts from the LQ image and ends at the desirable HQ image is novel. We show that the Markov chain can compress the predicted error by a factor of less than 1.
    \item We show that the image prior captured in a pretrained diffusion model can be harnessed by having the Markov chain built partially on the reverse diffusion process. Such a unique design also allows us to explicitly control the restoration's fidelity and realism by changing the Markov chain's length.
    \item We show that BFR can be achieved without complicated losses as in existing methods. Our method only needs to train a neural network with $L_1$ loss, simplifying the training pipeline. A preview of our results compared with existing methods is shown in Fig.~\ref{fig:method_visulization}.
\end{itemize}

\section{Related Work}
The BFR approaches mainly focused on exploring better face priors. We review a few popular priors in this section, including geometric priors, reference priors, and generative priors. Diffusion prior can be considered as a type of generative prior.

\noindent\textbf{Geometric Priors}. Face images are highly structured compared with natural images. The structural information, such as facial landmarks~\cite{chen2018fsrnet}, face parsing maps~\cite{chen2021progressive,shen2018deep,chen2021progressive}, facial component heatmaps~\cite{li2020blind}, and 3D shapes~\cite{ren2019face,hu2020face,zhu2022blind}, can be used a guidance to facilitate the restoration. The geometric face priors estimated from degraded inputs can be unreliable, affecting the performance of the subsequent BFR task.

\noindent\textbf{Reference Priors}. 
Some existing methods~\cite{li2018learning,dogan2019exemplar} guide the restoration with an additional HQ reference image that owns the same identity as the degraded input. The main limitations of these methods stem from their dependence on the HQ reference images, which are inaccessible in some scenarios.
Li~\textit{et al.}~\cite{li2020blind} address this limitation by constructing an offline facial component dictionary based on the features extracted from HQ images. Then, it searches for the closest facial components in this dictionary for the given LQ images during restoration.

\noindent\textbf{Generative Priors}. Unlike \cite{li2020blind},  more recent approaches directly exploit the rich priors encapsulated in generative models for BFR. Following the paradigm of GAN inversion~\cite{xia2022gan}, the earliest explorations~\cite{menon2020pulse,gu2020image,pan2021exploiting,mei2023ltt} iteratively optimize the latent code of a pretrained GAN for the desirable HQ target. To circumvent the time-consuming optimization, some studies~\cite{chan2021glean,wang2021towards,yang2021gan} directly embed the decoder of the pre-trained StyleGAN~\cite{karras2019style} into the BFR network, and evidently improve the restoration performance.
The success of VQGAN~\cite{esser2021taming} in image generation also inspires several BFR methods. These methods mainly design different strategies, e.g., cross-attention~\cite{wang2022restoreformer}, parallel decoder~\cite{gu2022vqfr}, and transformer~\cite{toward2022zhou}, to improve the matching between the codebook elements of the degraded input and the underlying HQ image.

Attributed to the powerful generation capability of the diffusion model, some works based on diffusion model have been proposed recently. Typically, SR3~\cite{saharia2022image} and SRDiff~\cite{li2022srdiff} both feed the LQ image into the diffusion model as a condition to guide the restoration in training. \zsyrevise{IDM~\cite{zhao2023towards} introduces an extrinsic pre-cleaning process to further improve the BFR performance on the basis of SR3.} To accelerate the inference speed, LDM~\cite{rombach2022high} proposed to train the diffusion model in the latent space of VQGAN~\cite{esser2021taming}.  These methods require one to retrain the diffusion model from scratch on some pre-collected data. Furthermore, the learned model would still be susceptible to the degradation mismatch when generalizing to other datasets. In a bid to circumvent the laborious and time-consuming retraining process, \zsyrevise{several investigations have explored the utilization of a pre-trained diffusion model as a generative prior to facilitate the restoration task, such as ILVR~\cite{Choi_2021_ICCV}, RePaint~\cite{Lugmayr_2022_CVPR}, DDRM~\cite{kawar2022denoising}, CCDF~\cite{chung2022come}, MCG~\cite{chung2022improving}, GDP~\cite{Fei_2023_CVPR}, DR2~\cite{wang2023dr2} and so on~\cite{wang2023zeroshot,zhu2023denoising,yue2023resshift,wang2023exploiting,lin2023diffbir}}. The common idea underlying these approaches is to modify the reverse sampling process of the pre-trained diffusion model by introducing a well-defined or manually assumed degradation model as an additional constraint. Even though these methods perform well in certain ideal scenarios, they can not deal with the BFR task since its degradation model is unknown and complicated. To address these issues, the present study devises a new learning paradigm for BFR based on a pre-trained diffusion model.

\zsyrevise{It is noteworthy that the concurrent study by Qiu \textit{et al.}~\cite{qiu2023diffbfr}, DiffBFR, similarly leverages a pre-trained diffusion model to facilitate the BFR task. It specifically designs an identity restoration module and a texture enhancement module to address the challenges posed by the long-tail distribution in BFR. This work represents a significant advancement in the diffusion-based BFR approaches.}

\section{Preliminaries}\label{sec:preliminary}
Before presenting our method, we commence with a brief introduction to the diffusion probabilistic model, denoted as diffusion model for conciseness. Diffusion model can be traced back to its seminal exposition in~\cite{sohl2015deep}, primarily inspired by the theory of non-equilibrium statistical physics. Ho et al.~\cite{ho2020denoising} reformulated this model from the perspective of denoising scoring matching, namely DDPM, achieving a huge success in the domain of image generation. In the course of this study, we follow the notations of DDPM for convenience. 

Diffusion model consists of a forward process (or diffusion process) and a reverse process. Given a data point $\bm{x}_0$ with probability distribution $q(\bm{x}_0)$, the forward process gradually destroys its data structure by repeated application of the following Markov diffusion kernel:
\begin{equation}
    q(\bm{x}_t|\bm{x_{t-1}}) =
    \mathcal{N}(\bm{x}_t;\sqrt{1-\beta_t}\bm{x}_{t-1}, \beta_t\bm{I}),
    \label{eq:diffused_kernel}
\end{equation}
where $t\in\{1,2,\cdots,T\}$, $\{\beta_t\}_{t=1}^T$ is a pre-defined or learned noise variance schedule, and $\bm{I}$ is the identity matrix. With a rational design on the variance schedule, it theoretically guarantees that $q(\bm{x}_T)$ converges to the unit spherical Gaussian distribution. It is noteworthy that the marginal distribution at arbitrary timestep $t$ has the following analytical form: 
\begin{equation}
    q(\bm{x}_t|\bm{x}_0) =
    \mathcal{N}(\bm{x}_t;\sqrt{\alpha_t}\bm{x}_0, (1-\alpha_t)\bm{I}),
    \label{eq:xt_cond_x0}
\end{equation}
where $\alpha_t = \prod_{l=1}^t (1-\beta_l)$.

As for the reverse process, it aims to learn a transition kernel from $\bm{x}_t$ to $\bm{x}_{t-1}$, which is defined as the following Gaussian distribution:
\begin{equation}
    p_{\theta}(\bm{x}_{t-1}|\bm{x}_t) =
    \mathcal{N}\left(\bm{x}_{t-1};\bm{\mu}_{\theta}(\bm{x}_t,t), \sigma^2_{t}\bm{I}\right),
    \label{eq:reverse_t_t1}
\end{equation}
where $\theta$ is the learnable parameter. With such a learned transition kernel, we can approximate the data distribution $q(\bm{x}_0)$ via the following marginal distribution:
\begin{equation}
    p_{\theta}(\bm{x}_0) = \int p(\bm{x}_T)\prod_{t=1}^T p_{\theta}(\bm{x}_{t-1}|\bm{x}_t)\mathrm{d}\bm{x}_{1:T},
    \label{eq:reverse_marginal}
\end{equation}
where $p(\bm{x}_T)=\mathcal{N}(\bm{x}_T;\bm{0},\bm{I})$.

In the framework of DDPM~\cite{ho2020denoising}, the reparametrization of the reverse posterior $p_{\theta}(\bm{x}_{t-1}|\bm{x}_t)$ is delineated as follows:
\begin{gather}
    \bm{\mu}_{\theta}(\bm{x}_t, t) = \sqrt{\frac{\alpha_{t-1}}{\alpha_{t}}}\left(
    \bm{x}_t - \frac{\beta_t}{\sqrt{1-\alpha_t}} \bm{\varepsilon}_{\theta}(\bm{x}_t, t)
    \right), \notag \\
    \sigma^2_t = \frac{1-\alpha_{t-1}}{1-\alpha_t}\beta_t,
    \label{eq:ddpm_reverse}
\end{gather}
where $\bm{\varepsilon}_{\theta}(\bm{x}_t, t)$ is designed to predict the noise contained in $\bm{x}_t$. Song et al.~\cite{song2021denoising} generalize DDPM to a non-Markov process named DDIM, resulting in a deterministic generative process. Within this paradigm, the reverse posterior is expressed as:
\begin{gather}
    \bm{\mu}_{\theta}(\bm{x}_t, t) = \sqrt{\alpha_{t-1}}\hat{\bm{x}}_0^{(t)} + \sqrt{1-\alpha_{t-1}-\sigma_t^2} \bm{\varepsilon}_{\theta}(\bm{x}_t, t), \notag \\
    \sigma^2_t = \eta \cdot \frac{1-\alpha_{t-1}}{1-\alpha_t}\beta_t, ~\eta \in [0,1],
    \label{eq:ddim_reverse}
\end{gather}
where 
\begin{equation}
    \hat{\bm{x}}_0^{(t)} = 
    \frac{\bm{x}_t - \sqrt{1-\alpha_t} \bm{\varepsilon}_{\theta}(\bm{x}_t, t)}{\sqrt{\alpha_t}}.
    \label{eq:x0_ddim}
\end{equation}
It is worth highlighting that the hyper-parameter $\eta$ controls the extent of randomness during inference. Specifically, when $\eta=0$, the reverse process becomes totally deterministic, while at $\eta=1$ it degenerates into the DDPM model.

\zsyrevise{The marginal distribution of Eq.~\eqref{eq:xt_cond_x0} indicates that the initial state $\bm{x}_0$ is contracted with a scaling factor $\sqrt{\alpha_t}$ after transitioning to timestep $t$. Imaging $\bm{x}_0$ as a coarsely estimated clean face image in BFR, this marginal distribution will naturally compress the predicted error associated with $\bm{x}_0$. Interestingly, recent work CCDF~\cite{chung2022come} also observes this error contraction property inherited from the diffusion model and successfully applies it to solve some inverse problems. Motivated by such an observation, this study aims to propose a more robust BFR method, thus advancing the efficacy of BFR techniques.}
\begin{figure*}[t]
    \centering
    \includegraphics[width=\linewidth]{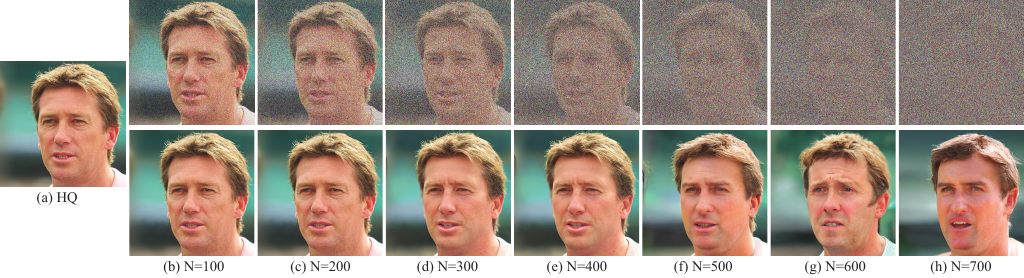}
    \caption{Illustration of the diffused $\bm{x}_N$ (top row) and the reconstructed results
        (bottom row) by a pretrained diffusion model from different
        starting timesteps. Note that the employed diffusion model is trained
        with 1000 discrete steps following~\cite{dhariwal2021diffusion}.}
    \label{fig:recons_comprison}
\end{figure*}
\begin{figure}[t]
    \centering
    \includegraphics[scale=0.36]{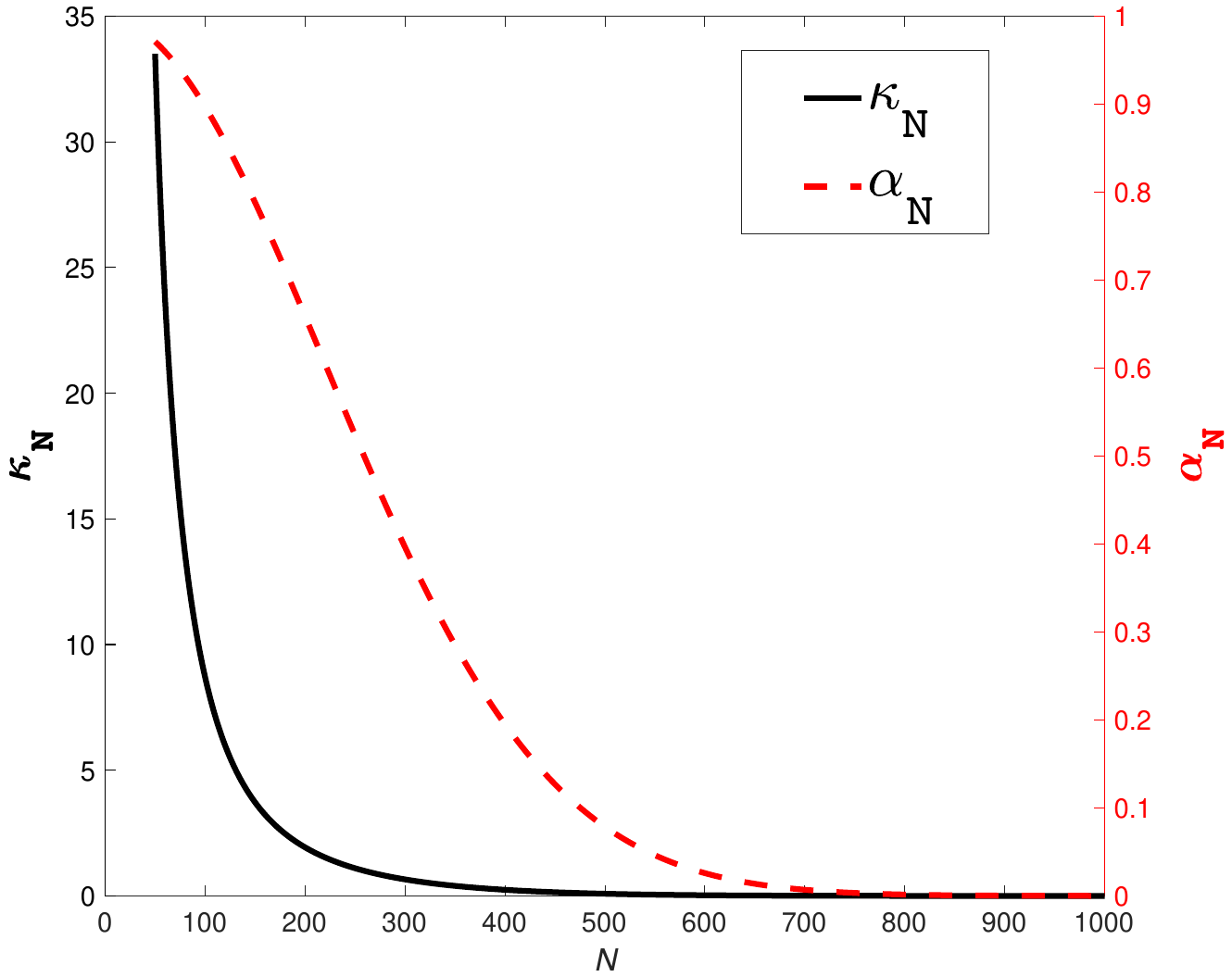}
    \vspace{-3mm}
    \caption{The curves of $\kappa_N$ and $\alpha_N$ with the starting timestep $N$.}
    \label{fig:curve_alpha}
\end{figure}

\section{Proposed Method} \label{sec:method}
In this section, we present the proposed BFR method \textit{DifFace} that exploits the image priors encapsulated in a pretrained diffusion model. To keep the notations consistent with Sec.~\ref{sec:preliminary}, we denote the LQ image and HQ image as $\bm{y}_0$ and $\bm{x}_0$. To restore the HQ image from its degraded counterpart, we aim at designing a rational posterior distribution of $p(\bm{x}_0|\bm{y}_0)$.

\subsection{Motivation} \label{subsec:motivation}
Considering a diffusion model with $T$ discrete steps, it provides a transition function from $\bm{x}_{t}$ to $\bm{x}_{t-1}$. With the aid of this transition, we can construct the posterior distribution $p(\bm{x}_0|\bm{y}_0)$ as follows:
\begin{equation}
    p(\bm{x}_0|\bm{y}_0) = \int p(\bm{x}_N|\bm{y}_0)\prod_{t=1}^N p_{\theta}(\bm{x}_{t-1}|\bm{x}_t)\mathrm{d}\bm{x}_{1:N},
    \label{eq:bfr_postirior}
\end{equation}
where $1 \le N < T$ is an arbitrary timestep. Therefore, we can restore $\bm{x}_0$ from $\bm{y}_0$ by sampling from this posterior using ancestral sampling~\cite{bishop2006pattern} from timestpe $N$ to $1$ as follows:
\begin{equation}
    \bm{x}_N \sim p(\bm{x}_N|\bm{y}_0), ~ ~
    \bm{x}_{t-1}|\bm{x}_t \sim p_{\theta}(\bm{x}_{t-1}|\bm{x}_t). 
    \label{eq:sampling_restore}
\end{equation}
Since the transition kernel $p_{\theta}(\bm{x}_{t-1}|\bm{x}_t)$ can be readily borrowed from a pretrained diffusion model, our goal thus turns to design the transition distribution of $p(\bm{x}_N|\bm{y}_0)$.

We have an important observation by delving into Eq.~(\ref{eq:sampling_restore}). If replacing $p(\bm{x}_N|\bm{y}_0)$ with the marginal distribution $q(\bm{x}_N|\bm{x}_0)$ defined in Eq.~(\ref{eq:xt_cond_x0}), Eq.~\eqref{eq:sampling_restore} degenerates into the diffusion and reconstruction process for $\bm{x}_0$ via a pretrained diffusion model, i.e.,
\begin{equation}
    \underbrace{\bm{x}_N \sim q(\bm{x}_N|\bm{x}_0)}_{\text{Diffusion}}, ~ ~
    \underbrace{\bm{x}_{t-1}|\bm{x}_t \sim p_{\theta}(\bm{x}_{t-1}|\bm{x}_t)}_{\text{Reconstruction}}.
    \label{eq:sampling_recons}
\end{equation}
In Figure~\ref{fig:recons_comprison}, we show some diffused and reconstructed results under different settings for the starting timestep $N$. One can observe that when $N$ lies in a reasonable range (e.g., $N<500$), meaning that $\bm{x}_0$ is slightly ``destroyed'', it is possible to accurately reconstruct $\bm{x}_0$ using the pretrained diffusion model.

This observation indicates that $q(\bm{x}_N|\bm{x}_0)$ is an ideal choice for the desired $p(\bm{x}_N|\bm{y}_0)$ by setting a reasonable $N$. Since the HQ image $\bm{x}_0$ is inaccessible in the task of BFR, we thus explore how to design a plausible $p(\bm{x}_N|\bm{y}_0)$ to approximate $q(\bm{x}_N|\bm{x}_0)$.

\begin{algorithm}[t]
    \caption{Inference procedure of \textit{DifFace}}
    \label{alg:difface-ddim0}
    \begin{algorithmic}[1]
        \REQUIRE ~~ \\
        The observed LQ image $\bm{y}_0$; \\
        The starting timestep $N$; \\
        \ENSURE ~~ \\
        The restored HQ image $\bm{x}_0$;
        \STATE Sample $\bm{x}_N$ from $p(\bm{x}_N|\bm{y}_0)$ according to Eq.~\eqref{eq:distribution_xs_y0}
        \FOR{$t=N, N-1, \cdots, 1$}
        \STATE Refine $\hat{\bm{x}}_0^{(t)}$ via Eq.~\eqref{eq:inpainting_solution} (Optional)
        \STATE $\bm{x}_{t-1} = \sqrt{\alpha_{t-1}}\hat{\bm{x}}_0^{(t)} + \sqrt{1-\alpha_{t-1}}\bm{\varepsilon}_{\theta}(\bm{x}_t, t) + \sigma_t \bm{\xi}$, where $\sigma_t$ is defined in Eq.~\eqref{eq:ddim_reverse}, $\bm{\xi} \sim \mathcal{N}(\bm{\xi}; \bm{0}, \bm{I})$
        \ENDFOR
    \end{algorithmic} 
\end{algorithm}
\subsection{Design}
Recall that our goal is to design a transition distribution $p(\bm{x}_N|\bm{y}_0)$ to approximate $q(\bm{x}_N|\bm{x}_0)$. Fortunately, the target distribution $q(\bm{x}_N|\bm{x}_0)$ has an analytical form as shown in Eq.~(\ref{eq:xt_cond_x0}). This inspires us to formulate $p(\bm{x}_N|\bm{y}_0)$ as a Gaussian distribution as follows:
\begin{equation}
    p(\bm{x}_N|\bm{y}_0) =
    \mathcal{N}\left(\bm{x}_N;\sqrt{\alpha_N}f(\bm{y}_0;w), (1-\alpha_N)\bm{I}\right),
    \label{eq:distribution_xs_y0}
\end{equation}
where $f(\cdot;w)$ is a neural network with parameter $w$, aiming to provide an initial prediction for $\bm{x}_0$. It should be noted that the final restored result by our method is achieved by sampling from the whole Markov chain of Eq.~(\ref{eq:sampling_restore}) (see Fig.~\ref{fig:framework}), and not directly predicted by $f(\cdot;w)$. As for $f(\cdot;w)$, it is only used to construct the marginal distribution of $\bm{x}_N$, a diffused version of $\bm{x}_0$, and thus named as ``diffused estimator'' in this work.

Next, we consider the Kullback-Leibler (KL) divergence between the designed $p(\bm{x}_N|\bm{y}_0)$ and its target $q(\bm{x}_N|\bm{x}_0)$. By denoting the predicted error of $f(\cdot;w)$ as $\bm{e}=\bm{x}_0 - f(\bm{y}_0;w)$, we have
\begin{equation}
    D_{\text{KL}} \left[p(\bm{x}_N|\bm{y}_0)\Vert q(\bm{x}_N|\bm{x}_0)\right] = 
    \frac{1}{2} \kappa_N \Vert \bm{e} \Vert_2^2,
    \label{eq:kl_qp}
\end{equation}
where $\kappa_N=\frac{\alpha_N}{1-\alpha_N}$. As shown in Fig.~\ref{fig:curve_alpha}, $\kappa_N$ strictly decreases monotonically with the timestep $N$. Hence, larger $N$ will offer a better approximation to $q(\bm{x}_N|\bm{x}_0)$, and further achieve a more realistic image via the designed posterior distribution of Eq.~\eqref{eq:bfr_postirior}. However, $\bm{x}_N$ will contain more noises when $N$ is getting larger, as shown in Fig.~\ref{fig:recons_comprison}. Thus, an overly large $N$ will inevitably deviate the restored result from the ground truth $\bm{x}_0$. Therefore, the choice of $N$ induces a realism-fidelity trade-off for the restored HQ image. We provide the ablation study in Sec.~\ref{subsec:exp_bft_analysis}.

\subsection{Algorithm} \label{subsec:algorithm}
Based on the constructed posterior distribution $p(\bm{x}_0|\bm{y}_0)$ of Eq.~\eqref{eq:bfr_postirior}, we can obtain the desired HQ image $\bm{x}_0$ through sequentially sampling from $p(\bm{x}_N|\bm{y}_0)$ and $p_{\theta}(\bm{x}_{t-1}|\bm{x}_t)$ for $t=N,\cdots,1$ as shown in Algorithm~\ref{alg:difface-ddim0}. However, it is essential to acknowledge that this stochastic sampling process inherently introduces potential deviation from the underlying HQ image, which is particularly problematic in the context of BFR that demands a high level of fidelity. To mitigate this fidelity issue, we utilize the DDIM sampler configured with $\eta=0.5$ when sampling from $p_{\theta}(\bm{x}_{t-1}|\bm{x}_t)$. This strategy effectively eliminates the inherent randomness in the sampling process to some extent, enhancing the fidelity of the reconstructed HQ image. Detailed ablation studies on the settings of $\eta$ can be found in Sec.~\ref{subsec:exp_bft_analysis}.

\begin{figure}
    \centering
    \includegraphics[width=\linewidth]{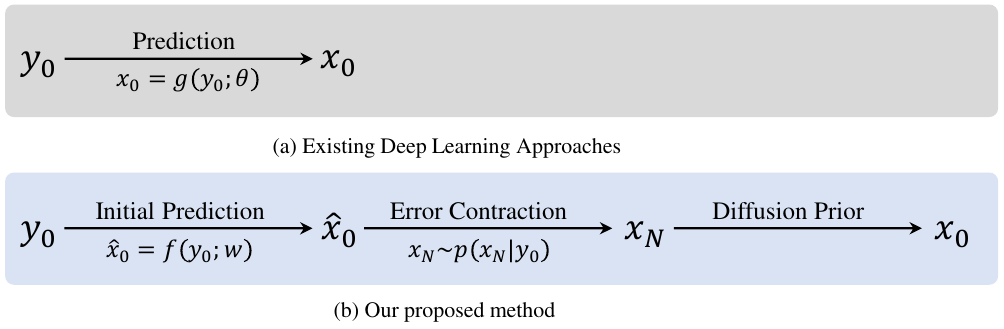}
    \vspace{-6mm}
    \caption{\zsyrevise{Comparison of existing deep learning-based approaches and our proposed method. For the former, we adopt $g(\cdot;\theta)$ to denote the learnable neural network with parameter $\theta$.}}
    \label{fig:comparison_framework}
\end{figure}
\subsection{Discussion} \label{subsec:remark}
\zsyrevise{Current deep learning-based BFR methods attempt to learn a straightforward mapping, namely $g(\cdot;\theta)$ in Fig.~\ref{fig:comparison_framework} (a), from the degraded LQ image $\bm{y}_0$ to the underlying HQ image $\bm{x}_0$ under multiple constraints. This strategy encounters formidable obstacles due to the inherent complexity of the degradation model in real-world scenarios. The proposed method circumvents this challenge by predicting $\bm{x}_N$, a diffused version of $\bm{x}_0$, from $\bm{y}_0$, as illustrated in Fig.~\ref{fig:comparison_framework} (b). Such a new learning paradigm brings several significant advantages over previous approaches:} 
\begin{itemize}[topsep=1pt,parsep=2pt,itemsep=3pt,leftmargin=15pt]
    \item \textbf{Error Contraction}. Considering the diffused estimator $f(\cdot;w)$, its predicted error is denoted as $\bm{e}=\bm{x}_0-f(\bm{y}_0;w)$. In our method, the problem of BFR is reformulated to predict $\bm{x}_N$ as explained in Sec.~\ref{subsec:motivation}. According to Eq.~\eqref{eq:distribution_xs_y0}, $\bm{x}_N$ can be accessed as follows:
        \begin{align}
            \bm{x}_N &= \sqrt{\alpha_N}f(\bm{y}_0;w) + \sqrt{(1-\alpha_N)}\bm{\zeta}  \notag \\
                     &= \sqrt{\alpha_N}\bm{x}_0 - \sqrt{\alpha_N}\bm{e} + \sqrt{(1-\alpha_N)}\bm{\zeta},
            \label{eq:sampling_xs_error1}
        \end{align}
        where $\bm{\zeta} \sim \mathcal{N}(\bm{\zeta}|\bm{0}, \bm{I})$.
        It can be easily seen that the predicted error $\bm{e}$ is contracted by a factor of $\sqrt{\alpha_N}$ after transitioning from $\bm{x}_0$ to $\bm{x}_N$,
        where $\sqrt{\alpha_N}$ is less than 1 as shown in Fig.~\ref{fig:curve_alpha}. We call such a characteristic induced by our method as ``error contraction''.

        \zsyrevise{Since the real-world degradation model is unknown, it is thus difficult to train an ideal restorer for BFR, e.g., $g(\cdot;\theta)$ or $f(\cdot;w)$ in Fig.~\ref{fig:comparison_framework}. However, our method has a larger error tolerance to $f(\cdot;w)$ attributed to the error contraction property. Intuitively, this appealing capability of compressing errors improves the robustness of our method, especially when dealing with severe and complex degradations. This is further validated in the following experiments of Sec.~\ref{sec:exp_bfr}. What's more, benefiting from the error contraction, the diffused estimator $f(\cdot;w)$ in our method can be simply trained with $L_1$ loss on some synthetic data, bypassing the sophisticated training process compared with most recent methods~\cite{li2020blind,wang2021towards,gu2022vqfr}.}
        
    \item \textbf{Diffusion Prior}. After obtaining the diffused $\bm{x}_N$ via Eq.~\eqref{eq:sampling_xs_error1}, our method gradually generates the desirable HQ result by sampling recursively from $p_{\theta}(\bm{x}_{t-1}|\bm{x}_t)$ starting from $t=N$ and ending at $t=1$. Through this sampling procedure, we effectively leverage the rich image priors and powerful generation capability of the pre-trained diffusion model to help the restoration task. Unlike existing methods, since the diffusion model is completely trained on the HQ images in an unsupervised manner, it thus reduces the dependence of our method on the manually synthesized training data, of which the distribution may deviate from the true degradation.
\end{itemize}

\textit{These two intrinsic properties are mainly delivered by the constructed posterior in Eq.~\eqref{eq:bfr_postirior}, which is the core formulation of our method. Such a posterior is specifically designed for BFR to mitigate the robustness issue caused by complicated degradations. Therefore, our proposed method indeed renders a new simple yet robust learning paradigm for BFR, rather than a straightforward application of diffusion model in this problem}.

\vspace{1mm}
\noindent\zsyrevise{{\textbf{Remark}}. It should be noted that recent work CCDF~\cite{chung2022come} also exploits the property of error contraction induced by the diffusion model in several inverse problems. Based on this property, it establishes a sound theorem that depicts the convergence error bound from the perspective of optimization. Our study parallels this appealing property, employing it to construct a simple yet precise posterior distribution for BFR. From the Bayesian inference standpoint, we quantify the statistical estimation error via KL divergence as shown in Eq.~\eqref{eq:kl_qp}. These findings are expected to inspire further investigations in relevant research areas.}

\section{Extension} \label{sec:extension}
The proposed \textit{DifFace} constitutes a versatile framework, being able to easily extended to a variety of blind image restoration tasks, such as super-resolution, deblurring and so on. The extensibility of \textit{DifFace} can be readily achieved through a straightforward substitution of the training data to train the diffused estimator. This section delves into a distinct facet of non-blind restoration, typified by a well-defined degradation model, e.g., inpainting and colorization. The availability of the degradation model in these scenarios enables us to incorporate an additional optimization strategy to further improve the efficacy of our method. For clarity and pedagogical purposes, we take the task of inpainting as an example to elucidate the precise technical details involved. 

The goal of inpainting is to restore the underlying HQ image $\bm{x}_0$ given the observed LQ image $\bm{y}_0$, while adhering to the following constraint, i.e., 
\begin{equation}
     \bm{y}_{0_{[i,j]}} = \begin{dcases}
         \bm{x}_{0_{[i,j]}}, & \text{if} ~ \bm{M}_{[i,j]}=0, \\
         0,             &\text{if} ~  \bm{M}_{[i,j]}=1,
     \end{dcases}
    \label{eq:inpainting_constraint}
\end{equation}
where $\bm{A}_{[i,j]}$ denotes the element located at position $(i,j)$ of the matrix $\bm{A}$, and $\bm{M}$ represents the binary image mask filled with $0$ or $1$. In the inference procedure of \textit{DifFace}, we generate $\bm{x}_{t-1}$ from $\bm{x}_t$ based on the estimated value for $\bm{x}_0$, namely $\hat{\bm{x}}_0^{(t)}$, in each timestep as exemplified in Eq.~\eqref{eq:ddim_reverse}. This inspires us to refine $\hat{\bm{x}}_0^{(t)}$ according to the degradation constraint of Eq.~\eqref{eq:inpainting_constraint}. By denoting the refined target as $\ddot{\bm{x}}_0^{(t)}$, the refinement process can be formulated as an optimization problem:
\begin{equation}
    \min_{\ddot{\bm{x}}_0^{(t)}} \left\Vert \ddot{\bm{x}}_0^{(t)} - \hat{\bm{x}}_0^{(t)} \right\Vert_2 + \gamma \left\Vert (\bm{1} - \bm{M}) \odot \left(\bm{y}_0 - \ddot{\bm{x}}_0^{(t)}\right) \right\Vert_2, 
    \label{eq:inparing_optimization}
\end{equation}
where $\odot$ denotes the Hadamard product, $\gamma$ is a hyper-parameter controlling the balance between the diffusion prior and the degradation constraint. Furthermore, this problem admits a closed-form solution as follows\footnote{All the mathematical operations in Eq~\eqref{eq:inpainting_solution} are pixel-wise.}:
\begin{equation}
    \ddot{\bm{x}}_0^{(t)} = \frac{\hat{\bm{x}}_0^{(t)} + \gamma(\bm{1}-\bm{M})^2\odot\bm{y}_0}{\bm{1} + \gamma(\bm{1}-\bm{M})^2}.
    \label{eq:inpainting_solution}
\end{equation}
It should be noted that in cases where a closed-form solution is not feasible,  Eq.~\eqref{eq:inpainting_solution} can be substituted with a single stochastic gradient descent (SGD) step.
\begin{figure*}[t]
    \centering
    \includegraphics[width=\linewidth]{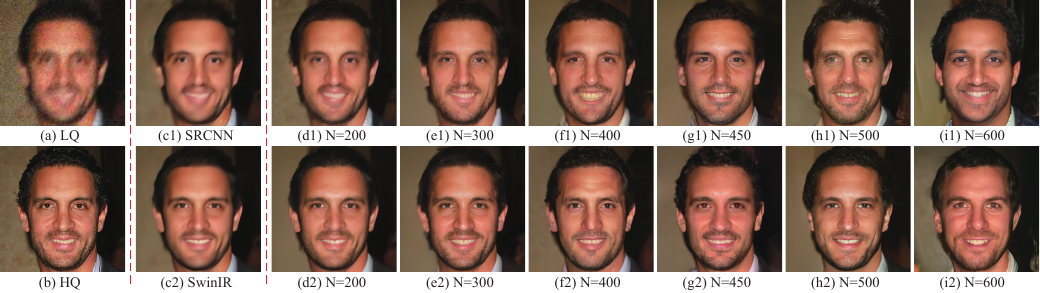}
    \vspace{-6mm}
    \caption{An example restored by \textit{DifFace} under different settings of the starting timestep $N$ and the diffused estimator $f(\cdot;w)$. (a) LQ image, (b) HQ image, (c1)-(c2) restored results by SRCNN and SwinIR, (d1)-(i1) restored results by \textit{DifFace} that takes SRCNN as a diffused estimator, (d2)-(i2) restored results by \textit{DifFace} that takes SwinIR as a diffused estimator.}
    \label{fig:time_restore}
    \vspace{-2mm}
\end{figure*}
\begin{table*}[t]
    \centering
    \caption{Quantitative comparisons of SRCNN, SwinIR and \textit{DifFace} on 
        the dataset of CelebA-Test for BFR. ``\textit{DifFace}\textcolor[gray]{0.5}{(X)}'' means that it takes the backbone ``\textcolor[gray]{0.5}{X}'' as diffused estimator. The parameters in ``\textit{DifFace}\textcolor[gray]{0.5}{(X)}'' includes both of that in the backbone ``\textcolor[gray]{0.5}{X}'' and the pre-trained diffusion model.}
        \label{tab:metirc_backbone}
    \small
    \vspace{-3mm}
    \begin{tabular}{@{}C{3.2cm}@{}|@{}C{1.7cm}@{} @{}C{1.7cm}@{} @{}C{1.7cm}@{}
                    @{}C{1.6cm}@{} @{}C{1.6cm}@{} @{}C{1.7cm}@{} @{}C{1.7cm}@{} @{}C{3.0cm}@{}}
        \Xhline{0.8pt}
        \multirow{2}*{Methods} & \multicolumn{8}{c}{Metrics} \\
        \Xcline{2-9}{0.4pt}
                        & PSNR$\uparrow$  & SSIM$\uparrow$     & LPIPS$\downarrow$   & IDS$\downarrow$   
                        & LMD$\downarrow$ & FID-F$\downarrow$  & FID-G$\downarrow$   & \# Parameters(M)     \\
        \Xhline{0.4pt}
        SRCNN           & 23.58           & 0.722           & 0.555              & 73.09         
                        & 11.13           & 133.49          & 108.68             & 1.03 \\
        \textit{DifFace}\textcolor[gray]{0.5}{(SRCNN)}
                        & 23.03           & 0.670           & 0.517              & 70.80            
                        & 8.40            & 58.53           & 44.83     & 1.03+\textcolor[gray]{0.5}{159.59} \\
        \hline\hline
        SwinIR          & 24.32           & 0.736           & 0.507              & 63.80           
                        & 5.82            & 131.82          & 93.73              & 15.79    \\
        \textit{DifFace}\textcolor[gray]{0.5}{(SwinIR)}
                        & 23.44           & 0.670           & 0.461              & 64.94            
                        & 6.06            & 48.98           & 20.29    &15.79+\textcolor[gray]{0.5}{159.59} \\
        \Xhline{0.8pt}
    \end{tabular}
\end{table*}
\begin{figure}[t]
    \centering
    \includegraphics[scale=0.35]{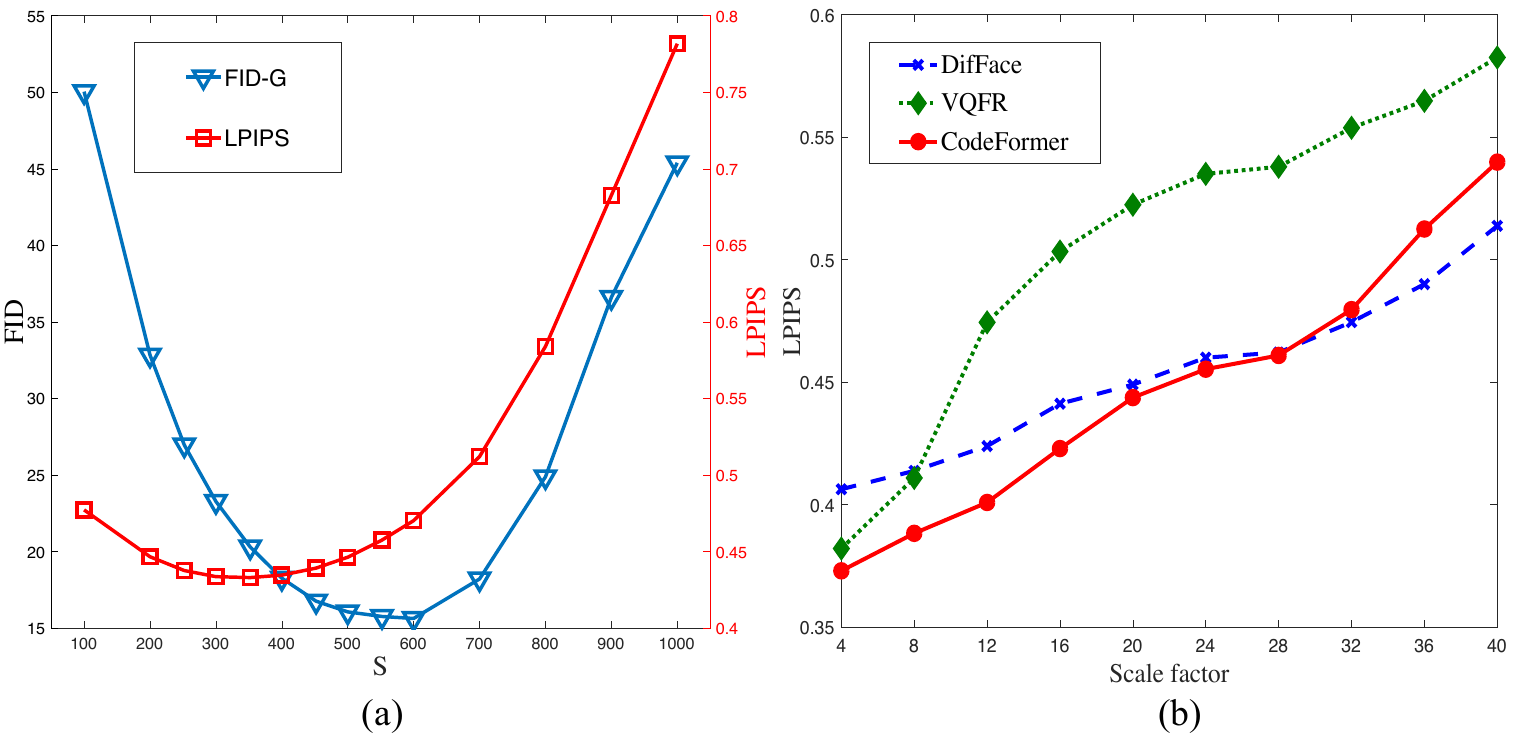}
    \vspace{-6mm}
    \caption{From left to right: (a) FID and LPIPS with respect to the starting timestep $N$ on the validation dataset, (b) LPIPS with respect to the downsampling factor, namely $s$ in Eq.~\eqref{eq:degradation_synthetic}, on the testing dataset.}
    \label{fig:fid_lpips}
\end{figure}

\section{Experimental results on BFR} \label{sec:exp_bfr}
In this section, we conduct extensive experiments to validate the effectiveness of the proposed \textit{DifFace} on the task of BFR. The empirical investigation commences with an exposition of the  experimental configurations. Subsequently, a thorough analysis is undertaken to assess the impact of three pivotal components, namely the starting timestep $N$, the diffused estimator $f(\cdot;w)$, and the hyper-parameter $\eta$ of DDIM in Eq.~\eqref{eq:ddim_reverse}, in our method. Finally, a comprehensive comparison to recent state-of-the-art (SotA) methods across both synthetic and real-world datasets is carried out to evaluate the overrall performance of \textit{DifFace}.

\subsection{Experimental setup} \label{subsec:exp_bfr_setup}
\noindent\textbf{Training Settings.} For the diffused estimator $f(\cdot;w)$, we considered two canonical network architectures as the backbones, i.e., SRCNN~\cite{dong2015image} and SwinIR~\cite{liang2021swinir}. They were both trained on the FFHQ dataset~\cite{karras2019style} that contains 70k HQ face images. We firstly resized the HQ images into a resolution of $512\times 512$, and then synthesized the LQ images following a typical degradation model used in recent literature~\cite{wang2021towards}:
\begin{equation}
    \bm{y} = \left\{\left[\left(\bm{x}*\bm{k}_l\right)\downarrow_s + \bm{n}_{\sigma}\right]_{\text{JPEG}_q}\right\}{\uparrow_s},
    \label{eq:degradation_synthetic}
\end{equation}
where $\bm{y}$ and $\bm{x}$ are the LQ and HQ image, $\bm{k}_l$ is the Gaussian kernel with kernel width $l$, $\bm{n}_{\sigma}$ is Gaussian noise with standard deviation $\sigma$, $*$ is 2D convolutional operator, $\downarrow_s$ and $\uparrow_s$ are the Bicubic downsampling or upsampling operators with scale $s$, and $[\cdot]_{\text{JPEG}_q}$ represents the JPEG compression process with quality factor $q$. And the hyper-parameters $l$, $s$, $\sigma$, and $q$ are uniformly sampled from $[0.1,15]$, $[0.8,32]$, $[0,20]$, and $[30,100]$ respectively. 
\begin{table}[t]
    \centering
    \caption{Performance of \textit{DifFace} under various configurations of the hyper-parameter $\eta$ on the dataset of CelebA-Test for BFR.}
        \label{tab:metirc_eta}
    \small
    \vspace{-3mm}
    \begin{tabular}{@{}C{1.4cm}@{}|@{}C{1.4cm}@{} @{}C{1.5cm}@{} @{}C{1.5cm}@{}
                                   @{}C{1.5cm}@{} @{}C{1.4cm}@{}}
        \Xhline{0.8pt}
        \multirow{2}*{Metrics} & \multicolumn{5}{c}{$\eta$} \\
        \Xcline{2-6}{0.4pt}
                           & 0       & 0.25      & 0.50      & 0.75          & 1.0       \\
        \Xhline{0.4pt}
        PSNR$\uparrow$     & 23.49   & 23.48     & 23.44     & 23.37         & 23.28      \\
        SSIM$\uparrow$     & 0.681   & 0.684     & 0.670     & 0.695         & 0.699      \\
        LPIPS$\downarrow$  & 0.475   & 0.471     & 0.461     & 0.449         & 0.443      \\
        IDS$\downarrow$    & 64.58   & 64.65     & 64.94     & 65.37         & 65.87      \\
        LMD$\downarrow$    & 6.018   & 6.030     & 6.056     & 6.091         & 6.132     \\
        FID-F$\downarrow$  & 51.81   & 50.68     & 48.98     & 48.32         & 50.29      \\
        FID-G$\downarrow$  & 23.51   & 22.18     & 20.29     & 19.04         & 19.63      \\
        \Xhline{0.8pt}
    \end{tabular}
\end{table}

We adopted the Adam~\cite{kingma2015adam} algorithm to optimize the network parameters $w$ under $L_1$ loss. The batch size was set as 16, and other settings of Adam followed the default configurations of Pytorch~\cite{paszke2019pytorch}. We trained the model for 500k iterations, and the learning rate was decayed gradually from $1e\text{-}4$ to $1e\text{-}6$ with the cosine annealing schedule~\cite{DBLP:conf/iclr/LoshchilovH17}. The leveraged diffusion model was trained on the FFHQ~\cite{karras2018progressive} dataset based on the official code\footnote{\url{https://github.com/openai/guided-diffusion}} of~\cite{dhariwal2021diffusion}. It contains 1,000 discrete diffusion steps, and is accelerated to 250 steps using the technique in~\cite{nichol2021improved} when applied in~\textit{DifFace}.

\vspace{1mm}\noindent\textbf{Testing Datasets}. We evaluate \textit{DifFace} on one synthetic dataset and three real-world datasets. The synthetic dataset, denoted as CelebA-Test, contains 4000 HQ images from CelebA-HQ~\cite{karras2018progressive}, and the corresponding LQ images are synthesized via Eq.~(\ref{eq:degradation_synthetic}). The specific settings on the hyper-parameters of the degradation can be found in the Appendix.
As for the real-world datasets, we consider three typical ones with different degrees of degradation, namely LFW, WebPhoto~\cite{wang2021towards}, and WIDER~\cite{toward2022zhou}. LFW consists of 1711 mildly degraded face images in the wild, which contains one image for each person in LFW dataset~\cite{huang2008labeled}. WebPhoto is made up of 407 face images crawled from the internet. Some of them are old photos with severe degradation. WIDER selects 970 face images with very heavy degradations from the WIDER Face dataset~\cite{yang2016wider}, thus is suitable to test the robustness of different methods under severe degradations.

\vspace{1mm}\noindent\textbf{Comparison Methods.} \zsyrevise{We compare \textit{DifFace} with seven recent BFR methods, including PULSE~\cite{menon2020pulse},  PSFRGAN~\cite{chen2021progressive}, GFPGAN~\cite{wang2021towards}, RestoreFormer~\cite{wang2022restoreformer}, VQFR~\cite{gu2022vqfr}, CodeFormer~\cite{toward2022zhou}, and DR2~\cite{wang2023dr2}. For these comparison methods, we all adopt the default configurations in their official code.}

\vspace{1mm}\noindent\textbf{Evaluation Metrics.} In order to comprehensively assess various methods, this study adopts six quantitative metrics following the setting of VQFR~\cite{gu2022vqfr}, namely PSNR, SSIM~\cite{wang2004image}, LPIPS~\cite{zhang2018unreasonable}, identity score (IDS), landmark distance (LMD), and FID~\cite{heusel2017gans}. Note that IDS, \zsyrevise{also referred to as ``Deg" in certain literature~\cite{gu2022vqfr}}, and LMD both serve as quantifiers for the identity between the restored images and their ground truths. IDS gauges the embedding angle of ArcFace~\cite{deng2019arcface}, while LMD calculates the landmark distance using $L_2$ norm between pairs of images. FID quantifies the KL divergence between the feature distributions, assumed as Gaussian distribution, of the restored images and a high-quality reference dataset. For the reference dataset, we employ both of the ground truth images and the FFHQ~\cite{karras2018progressive} dataset. The corresponding results computed under these two settings are denoted as ``FID-G'' and ``FID-F'' for clarity.
\begin{figure*}[t]
    \centering
    \includegraphics[scale=1.02]{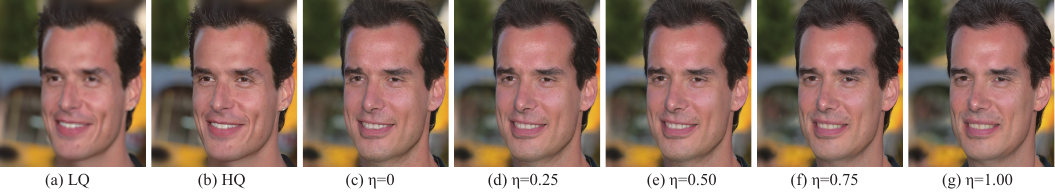}
    \vspace{-6mm}
    \caption{Visual comparisons of \textit{DifFace} across various configurations of the hyper-parameter $\eta$. Please zoom in for a better view.} \label{fig:ddim_eta}
    \vspace{-3mm}
\end{figure*}
\begin{table*}[t]
    \centering
    \caption{\zsyrevise{Quantitative comparisons of different BFR methods on the testing dataset of CelebA-Test. The best and second best results are highlighted in \textbf{bold} and \underline{underline}, respectively.}}
    \vspace{-2mm}
    \label{tab:metirc_celeba}
    \small
    \begin{tabular}{@{}C{1.4cm}@{}|
                    @{}C{2.1cm}@{} @{}C{2.2cm}@{} @{}C{2.3cm}@{}
                    @{}C{2.6cm}@{} @{}C{2.0cm}@{} @{}C{2.4cm}@{}
                    @{}C{1.6cm}@{} @{}C{1.5cm}@{}}
        \Xhline{0.8pt}
        \multirow{2}*{Metrics}  & \multicolumn{8}{c}{Methods} \\
        \Xcline{2-9}{0.4pt} &  PULSE~\cite{menon2020pulse}     
                            & PSFRGAN~\cite{chen2021progressive}    & GFPGAN~\cite{wang2021towards} 
                            & RestoreFormer~\cite{wang2022restoreformer} & VQFR~\cite{gu2022vqfr}   
                            & CodeFormer~\cite{toward2022zhou}  
                            & DR2~\cite{wang2023dr2}
                            & \textit{DifFace} \\
        \Xhline{0.4pt}                                                                                       PSNR$\uparrow$                 & 21.59              & 21.80             & 21.25         & 21.76          & 21.11                 & \underline{22.70}  & 21.69 & \textbf{23.44} \\
        SSIM$\uparrow$  & \underline{0.675}     & 0.615          & 0.615         & 0.589          & 0.570                 & 0.644   & 0.661           & \textbf{0.690} \\
        LPIPS$\downarrow$   & 0.519              & 0.522             & 0.532         & 0.518          & 0.506                 & \textbf{0.438} & 0.490    & \underline{0.461} \\
        IDS$\downarrow$     & 76.70                 & 73.84             & 73.49         & 71.26          & 72.23                 & \textbf{64.64}  & 72.49      & \underline{64.94} \\
        LMD$\downarrow$     & 9.91                  & 9.19              & 12.26         & 11.14          & 11.32                 & \underline{8.01}  & 8.69   & \textbf{6.06} \\
        FID-F$\downarrow$   & 71.07                 & 69.30             & 58.40         & 55.84          & \underline{52.70}     & 61.31   & 64.52   & \textbf{48.98} \\
        FID-G$\downarrow$   & 50.54                 & 68.19             & 68.61         & 58.64          & 62.46                 & \underline{25.99}  & 43.11 & \textbf{20.29} \\
        
        \Xhline{0.8pt}
    \end{tabular}
    \vspace{-3mm}
\end{table*}
\begin{figure*}[t]
    \centering
    \includegraphics[width=\linewidth]{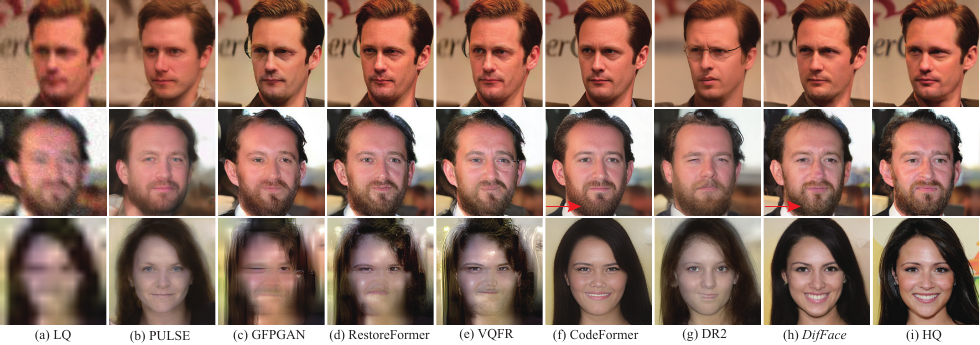}
    \vspace{-6mm}
    \caption{\zsyrevise{Visual comparisons of different methods on the synthetic dataset of CelebA-Test for BFR. Please zoom in for a better view.}}
    \label{fig:syn}
\end{figure*}

\subsection{Model Analysis} \label{subsec:exp_bft_analysis}
\vspace{1mm}\noindent\textbf{Starting Timestep}. In Fig.~\ref{fig:time_restore}, we show some restored results by \textit{DifFace} under different settings for the starting timesteps $N$.
The results show that one can make a trade-off between realism and fidelity through different choices of $N$. In particular, if we set a larger $N$, the restored results would appear more realistic but with lower fidelity in comparison to the ground truth HQ images. The phenomenon is reflected by the average FID~\cite{heusel2017gans} and LPIPS~\cite{zhang2018unreasonable} scores with respect to $N$ in Fig.~\ref{fig:fid_lpips}-(a).
The proposed \textit{DifFace} performs very well in the range of $[400, 450]$, and we thus set $N$ as $400$ throughout the whole experiments in this work. In practice, we speed up the inference four times following~\cite{nichol2021improved}, and thus sample 100 steps for each testing image.

\vspace{1mm}\noindent\textbf{Diffused Estimator}. Figure~\ref{fig:time_restore} displays an example restored by \textit{DifFace}, which either takes SRCNN~\cite{dong2015image} or SwinIR~\cite{liang2021swinir} as the diffused estimator $f(\cdot;w)$. The quantitative comparison results on these two backbones are listed in Table~\ref{tab:metirc_backbone}, and more details on the network architecture are provided in the Appendix. Even with the simplest SRCNN that only contains several plain convolutional layers, \textit{DifFace} is able to restore a plausible HQ image. Using a more elaborated architecture like SwinIR results in more apparent details (e.g., hairs). The results suggest the versatility of \textit{DifFace} in the choices of the diffused estimators. In the following experiments, we use SwinIR as our diffused estimator. 

\vspace{1mm}\noindent\textbf{Hyper-parameter $\eta$ of DDIM.} As expounded in Sec.~\ref{subsec:algorithm}, the hyper-parameter $\eta$ of Eq.~\eqref{eq:ddim_reverse} plays a crucial role in regulating the randomness of the inference process. In other words, the adjustment of the hyper-parameter $\eta$ offers a lever by which we can control the capability of fidelity preservation of \textit{DifFace}. A visual comparative example is presented in Fig.~\ref{fig:ddim_eta}, illustrating the restored outputs of \textit{DifFace} across varying values of $\eta$. This visual comparison clearly demonstrates the progressive deviation of the restored results from the underlying HQ image, as $\eta$ incrementally transitions from zero to one.

Furthermore, we provide a a summarized tabulation of the associated qualitative assessments in Table~\ref{tab:metirc_eta}. It is discernible that the decrease of $\eta$ yields favorable outcomes with respect to the fidelity-oriented metrics, specifically PSNR, IDS, and LMD. Conversely, this adjustment exerts an adverse influence on the perceptual metrics, including SSIM, LPIPS, and FID. Considering the stringent requirement on fidelity within the domain of BFR, we consistently set $\eta$ to be 0.5 throughout all experiments in this study.

\begin{table*}[t]
    \centering
    \caption{\zsyrevise{FID scores of different methods on three real-world testing datasets. The best and second best results are highlighted in \textbf{bold} and \underline{underline}, respectively.}}
    \label{tab:metirc_fid_real}
    \vspace{-3mm}
    \small
    \begin{tabular}{@{}C{1.5cm}@{}| @{}C{1.3cm}@{}|
                    @{}C{1.8cm}@{} @{}C{2.2cm}@{} @{}C{2.1cm}@{}
                    @{}C{2.5cm}@{} @{}C{1.7cm}@{} @{}C{2.2cm}@{}
                    @{}C{1.7cm}@{} @{}C{1.1cm}@{}}             
        \Xhline{0.8pt}
        \multirow{2}*{Datasets} & \multirow{2}*{\#Images} & \multicolumn{8}{c}{Methods} \\
        \Xcline{3-10}{0.4pt}
                          &     & PULSE~\cite{menon2020pulse}    & PSFRGAN~\cite{chen2021progressive}
                          & GFPGAN~\cite{wang2021towards}    &RestoreFormer~\cite{wang2022restoreformer}
                          & VQFR~\cite{gu2022vqfr}   & CodeFormer~\cite{toward2022zhou}
                          & DR2~\cite{wang2023dr2}   & \textit{DifFace} \\
        \Xhline{0.4pt}
        WIDER               & 970        & 69.59       & 49.85        & 39.46        &49.85        & 44.14         & \underline{38.85}  & 47.48     & \textbf{37.52} \\
        LFW               & 1711    & 65.30       & 49.64       & 50.03      & 48.50       & 50.78       & 52.43   & \underline{47.93}     & \textbf{46.80} \\
        WebPhoto          & 407       & 86.05       & 85.03       & 87.57       & \underline{77.36}      & \textbf{75.38}      & 83.27   & 108.81     & 81.60 \\
        \Xhline{0.8pt}
    \end{tabular}
    \vspace{-3mm}
\end{table*}
\begin{table*}[t]
    \centering
    \caption{\zsyrevise{Non-reference metrics of different methods on the real-world dataset WIDER. ``\textcolor[gray]{0.5}{Diffusion}'' denotes the average results on 3,000 images randomly generated by the pre-trained diffusion model, which can be regarded as the upper bound of \textit{DifFace}. }}
    \label{tab:non_reference_metirc}
    \vspace{-2mm}
    \small
    \begin{tabular}{@{}C{1.2cm}@{}| 
                    @{}C{1.9cm}@{} @{}C{2.1cm}@{} @{}C{2.1cm}@{} @{}C{2.5cm}@{} 
                    @{}C{1.8cm}@{} @{}C{2.3cm}@{} @{}C{1.6cm}@{} @{}C{1.2cm}@{} @{}C{1.4cm}@{}} 
        \Xhline{0.8pt}
        \multirow{2}*{Metrics} & \multicolumn{9}{c}{Methods} \\
        \Xcline{2-10}{0.4pt}
                         & PULSE~\cite{menon2020pulse}    & PSFRGAN~\cite{chen2021progressive} 
                         & GFPGAN~\cite{wang2021towards}  & RestoreFormer~\cite{wang2022restoreformer} 
                     & VQFR~\cite{gu2022vqfr}  &CodeFormer~\cite{toward2022zhou}  &DR2~\cite{wang2023dr2}
                     & \textit{DifFace} & \textcolor[gray]{0.5}{Diffusion }\\
        \Xhline{0.4pt}
        NIQE$\downarrow$ & 5.27     & 3.89     & 3.81  & 3.88  & 3.02  & 4.12 & 4.99  & 4.24             & \textcolor[gray]{0.5}{4.11} \\
        NRQM$\uparrow$   & 4.05     & 8.01    & 8.07   & 8.54  & 8.78  & 8.49 & 5.77  & 6.11             & \textcolor[gray]{0.5}{6.60} \\
        PI$\downarrow$   & 5.88     & 3.33    & 4.45   & 2.84  & 2.17  & 3.01  & 5.13  & 4.46             & \textcolor[gray]{0.5}{4.17} \\
        \Xhline{0.8pt}
    \end{tabular}
    \vspace{-2mm}
\end{table*}
\begin{figure*}[t]
    \centering
    \includegraphics[width=\linewidth]{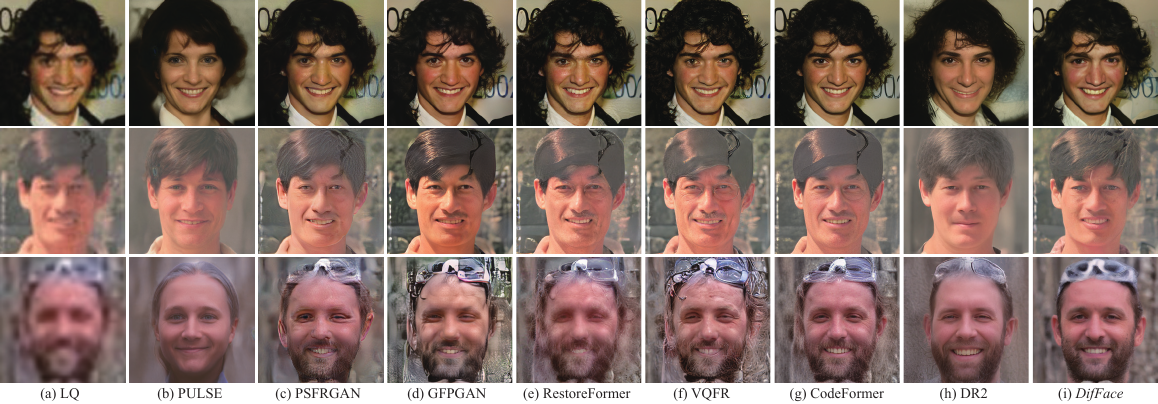}
    \vspace{-6mm}
    \caption{\zsyrevise{Visual comparisons of different methods on the real-world examples from LFW (first row), WebPhoto (second row), and WIDER (third row).}}
    \label{fig:real}
\end{figure*}
\subsection{Evaluation on Synthetic Data}
We summarize the comparative results on CelebA-Test in Table~\ref{tab:metirc_celeba}. The proposed \textit{DifFace} achieves the best or second-best performance across all five metrics, indicating its effectiveness and superiority on the task of BFR.
To further verify the robustness of \textit{DifFace}, we compare it with recent state-of-the-art methods, namely CodeFormer and VQFR, using LPIPS under different degrees of degradation in Fig.~\ref{fig:fid_lpips}-(b). We generate different degrees of degradation by increasing the scale factors from 4 to 40 with step 4 gradually, to the images in CelebA-Test. A total of 400 image pairs are generated for each scale. Figure~\ref{fig:fid_lpips}-(b) records the averaged performance of different methods over these 400 images with respect to different downsampling scale factors. While \textit{DifFace} is slightly inferior to CodeFormer and VQFR under small scale factors (mild degradations), its performance drops more gracefully and surpasses them in the cases of larger factors (severe degradations). These results substantiate the robustness of our \textit{DifFace} in the scenarios with very severe degradations and is consistent with the analysis in Sec.~\ref{subsec:remark}.

For visualization, three typical examples of the CelebA-Test are shown in Fig.~\ref{fig:syn}. In the first example with mild degradation, most of the methods are able to restore a realistic-looking image. PULSE produces a plausible face but fails to preserve the identity since its optimization cannot find the correct latent code through GAN inversion.
In the second and third examples that exhibit more severe degradation, only CodeFormer and \textit{DifFace} can handle such cases and return satisfactory face images. However, the results of CodeFormer still contain some slight artifacts in the areas of whiskers (marked by red arrows in Fig.~\ref{fig:syn}). Compared with CodeFormer, \textit{DifFace} performs more stably under this challenging degradation setting. Such robustness complies with our observation in Fig.~\ref{fig:fid_lpips}-(b).

\subsection{Evaluation on Real-world Data}\label{sec:exp_real}
In the experiments on real-world datasets, we mainly adopt FID-F as the quantitative metric since the ground truths are inaccessible. Specifically, FID-F calculates the KL divergence between the estimated feature statistics of the restored images and the FFHQ~\cite{karras2019style} reference dataset. The comparative results are summarized in Table~\ref{tab:metirc_fid_real}. We can observe that \textit{DifFace} achieves the best performance on both WIDER and LFW. On the WebPhoto, it also surpasses most recent BFR methods. It should be noted that the FID values on WebPhoto may not be representative as this dataset contains too few images (total 407) to estimate the feature distribution of the restored images. To supplement the analysis, we show three typical examples of these datasets in Fig.~\ref{fig:real}, and more visual results are given in Fig.~\ref{fig:real_appedix} of Appendix. It is observed that all the comparison approaches perform well in the first example with slight degradation, while \textit{DifFace} provides significantly better results in the second and third examples where the LQ images are severely degraded. This stable performance of \textit{DifFace} consistent with the robustness analysis in Fig.~\ref{fig:fid_lpips}-(b), mainly owing to the inherent error contraction mechanism. 

\zsyrevise{In addition, we did investigate several non-reference metrics, including NIQE~\cite{mittal2012making}, NRQM~\cite{ma2017learning}, and PI~\cite{blau20182018}. Table~\ref{tab:non_reference_metirc} summarizes the detailed comparison on the WIDER dataset, one of the most challenging benchmarks in BFR. Despite the notable superiority of \textit{DifFace} as validated above, it exhibits surprisingly weak performance against GAN-based methods across these non-reference metrics.
To explore the underlying reasons for this discrepancy, we randomly generated 3,000 HQ face images using the pre-trained diffusion model and evaluated their image quality using the aforementioned non-reference metrics. The average results, denoted as ``Diffusion'' and highlighted in \textcolor[gray]{0.5}{gray} in Table~\ref{tab:non_reference_metirc}, can be regarded as the upper bound performance of \textit{DifFace}. Interestingly, this upper bound still falls short of the performance achieved by GAN-based restoration methods. We guess that these non-reference metrics favor sharp images, even if they contain some artifacts, whereas the $L_2$ loss used to train the diffusion model tends to produce smooth results, which may not align with the preferences of these metrics. Further investigation is necessary to understand the performance gap introduced by these non-reference metrics.}

\subsection{Multiple Predictions}
\begin{figure*}[t]
    \centering
    \includegraphics[width=\linewidth]{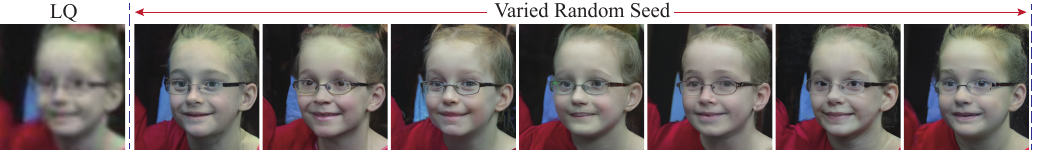}
    \vspace{-6mm}
    \caption{Unlike most existing BFR methods, \textit{DifFace} can generate multiple
        diverse and plausible results given a LQ image, by setting different random seeds for
        the diffusion model. This example is extracted from real-world dataset WIDER-Test.}
    \label{fig:random_real}
\end{figure*}

Most existing BFR methods produce only one HQ image for each LQ input, although there may be many reasonable possibilities. This is because they only learn a deterministic mapping between the LQ and HQ images. It is interesting to note that \textit{DifFace}, as shown in Fig.~\ref{fig:random_real}, is capable of producing multiple diverse and plausible HQ solutions for any given LQ image by setting different random seeds for the pre-trained diffusion model. More results are shown in Fig.~\ref{fig:random_real_appedix} of Appendix. This ``pluralistic'' property is favorable in BFR, as there exist many different HQ images that can generate the same LQ counterpart.
\begin{table*}[!t]
    \centering
    \caption{Quantitative comparisons of various methods on the testing dataset of inpainting. The best and second best results are highlighted in \textbf{bold} and \underline{underline}, respectively.}
    \label{tab:metric_inpainting}
    \vspace{-3mm}
    \small
    \begin{tabular}{@{}C{1.8cm}@{}| @{}C{1.3cm}@{}|
                    @{}C{2.2cm}@{} @{}C{1.7cm}@{} @{}C{1.6cm}@{} @{}C{2.0cm}@{}
                    @{}C{1.9cm}@{} @{}C{2.4cm}@{} @{}C{1.7cm}@{} @{}C{1.5cm}@{}}
        \Xhline{0.8pt}
         \multirow{2}*{Mask Types}  &\multirow{2}*{Metrics} & \multicolumn{8}{c}{Methods} \\
         \Xcline{3-10}{0.4pt}
              & & DeepFillv2~\cite{yu2019free}  &DSI~\cite{peng2021generating}   & LaMa~\cite{suvorov2022resolution}   & RePaint~\cite{Lugmayr_2022_CVPR}   & DDRM~\cite{kawar2022denoising}   & Score-SDE~\cite{song2020score}   & MCG~\cite{chung2022improving}   & \textit{DifFace} \\
        \Xhline{0.4pt}
        \multirow{3}*{Box} & LPIPS$\downarrow$  & 0.101  & 0.100  & \textbf{0.082}   & 0.108   & 0.137  & 0.147  & 0.119  & \underline{0.090} \\
            & FID-F$\downarrow$  & 78.95  & 77.71  & 85.10   & 92.33    & 99.92  & \textbf{64.44}   & \underline{73.99}  & 75.22 \\
            & FID-G$\downarrow$  & 15.00  & 14.20  & \textbf{12.31}   & 16.41    & 22.62  & 22.78   & 15.95  & \underline{12.35} \\
        \hline 
        \multirow{3}*{Irregular} & LPIPS$\downarrow$  & 0.230  & 0.227  & \textbf{0.184}   & 0.221   & 0.242  & 0.289  & 0.220  & \underline{0.187} \\
            & FID-F$\downarrow$  & 82.54  & 77.60  & 86.88   & 92.87    & 102.73  & \textbf{59.78}   & \underline{68.17}  & 71.14 \\
            & FID-G$\downarrow$  & 41.04  & 34.34  & \underline{25.79}   & 27.51    & 33.56  & 47.50   & 28.09  & \textbf{22.90} \\
            \hline 
        \multirow{3}*{Half} & LPIPS$\downarrow$  & 0.270  & 0.271  & \underline{0.224}   & 0.253   & 0.280  & 0.304  & 0.254  & \textbf{0.210} \\
            & FID-F$\downarrow$  & 109.86  & 87.95  & 95.27   & 96.85    & 109.93  & \textbf{61.30}   & \underline{71.71}  & 77.58 \\
            & FID-G$\downarrow$  & 74.86  & 48.26  & 32.07   & \underline{30.83}   & 38.85  & 49.42   & 32.94  & \textbf{24.97} \\
             \hline 
        \multirow{3}*{Expand} & LPIPS$\downarrow$  & 0.475  & 0.465  & \underline{0.403}   & 0.456   & 0.475  & 0.517  & 0.434  & \textbf{0.360} \\
            & FID-F$\downarrow$  & 190.08  & 112.88  & 131.21   & 109.12    & 138.43  & \textbf{65.49}   & \underline{71.98}  & 91.61 \\
            & FID-G$\downarrow$  & 162.07  & 91.87  & 68.75   & 45.25   & 61.07  & 77.06   & \underline{44.46}  & \textbf{37.59} \\
            \hline
        \multirow{3}*{Average} & LPIPS$\downarrow$  & 0.269  & 0.266  & \underline{0.223}   & 0.260   & 0.284  & 0.314  & 0.257  & \textbf{0.212} \\
            & FID-F$\downarrow$  & 115.36  & 89.03  & 99.61   & 97.79    & 112.75  & \textbf{62.75}   & \underline{71.46}  & 78.89 \\
            & FID-G$\downarrow$  & 73.24  & 47.17  & 34.73   & \underline{30.00}   & 39.03  & 49.19   & 30.36  & \textbf{24.45} \\
        \Xhline{0.8pt}
    \end{tabular}
\end{table*}
\begin{figure*}[t]
    \centering
    \includegraphics[width=\linewidth]{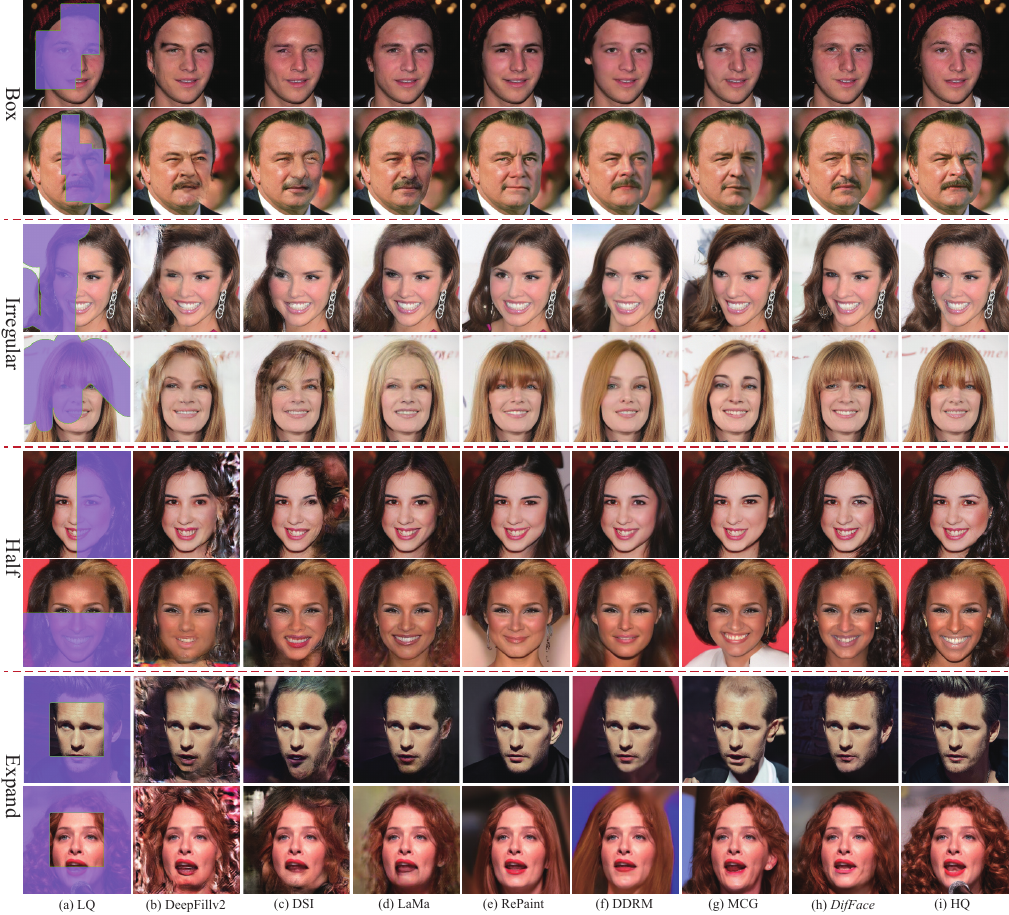}
    \vspace{-6mm}
    \caption{Visual comparisons of various methods on the testing dataset of inpainting. The masked areas are highlighted using a purple color. Please zoom in for a better view.}
    \label{fig:syn_inpainting}
\end{figure*}

\section{Experimental results on Inpainting}
\subsection{Experimental Setup}
\noindent\textbf{Training Settings.} We directly borrowed the network architecture introduced in a recent work, LaMa~\cite{suvorov2022resolution}, to serve as the diffused estimator for inpainting. Our training data consisted of 70k face images with a resolution of $256 \times 256$, all sourced from the FFHQ~\cite{karras2018progressive} dataset. For the generation of image masks, we randomly synthesize them during training, adhering to the configurations of LaMa~\cite{suvorov2022resolution}. In our model training, the Adam~\cite{kingma2015adam} algorithm is adopted with a batch size of 64. We conducted a total of 500k iterations, optimizing the model under the supervision of the $L_1$ loss function. The learning rate was progressively reduced from $1e\text{-}4$ to $5e\text{-}6$ in accordance with the cosine annealing schedule~\cite{DBLP:conf/iclr/LoshchilovH17}. 

\vspace{1mm}\noindent\textbf{Testing Dataset:} We randomly select 2000 images of size $256\times 256$ from CelebA-HQ~\cite{karras2018progressive} as our testing dataset. In order to ensure a thorough evaluation, we considered four distinct mask types, encompassing ``Box'' masks, ``Irregular'' masks, ``Half'' masks, and ``Expand'' masks, as visually shown in Fig.~\ref{fig:syn_inpainting}. For each category, we initially generate 500 random masks, and subsequently synthesize the LQ image based on these respective masks.

\vspace{1mm}\noindent\textbf{Comparison Methods:} In our comparative analysis, we evaluate the performance of \textit{DifFace} against three GAN-based methods, including DeepFillv2~\cite{yu2019free}, DSI~\cite{peng2021generating}, and LaMa~\cite{suvorov2022resolution}, as well as four diffusion-based methods, namely RePaint~\cite{Lugmayr_2022_CVPR}, DDRM~\cite{kawar2022denoising}, Score-SDE~\cite{song2020score}, and MCG~\cite{chung2022improving}.

\vspace{1mm}\noindent\textbf{Evaluation Metrics.} To assess the efficacy of various approaches, we adopt the metrics of LPIPS~\cite{zhang2018unreasonable} and FID~\cite{heusel2017gans} following LaMa~\cite{suvorov2022resolution}. Similarly as that in BFR, we provide both FID-F and FID-G scores for comprehensive reference.

\subsection{Comparison with SotA methods}
We present a quantitative comparison to various approaches on the testing dataset, as detailed in Table~\ref{tab:metric_inpainting}. The following key observations can be discerned from the tabulated results:
\begin{itemize}
    \item \textbf{Overall Superiority.} The proposed \textit{DifFace} attains either remarkable or, at the very least, comparable performance across most evaluation scenarios. This is especially  evident when scrutinizing the averaged results over four different mask types, thereby underscoring its efficacy and preeminence in the field of inpainting. 
    \item \textbf{Robustness to Large Occlusions.} Particularly noteworthy is the notable superiority exhibited by \textit{DifFace} when confronted with the ``Half'' and ``Expand'' masks with large occlusions, \textit{DifFace} markedly outperforms its counterparts. This pronounced performance shows the resilience of our method to severe degradation.
    \item \textbf{Necessity of Diffusion Prior.} The diffusion-based methods consistently outshine their GAN-based counterparts, indicating the indispensability of incorporating the diffusion prior in this task.
\end{itemize}
These findings collectively shed light on the promising potential and superiority of \textit{DifFace} for  face inpainting.

\begin{table*}[t]
    \centering
    \caption{Comparisons of different methods with respect to running time (s) on the task of inpainting. Note that the running time is counted on the NVIDIA Tesla V100 with a resolution of $256\times 256$.}
    \label{tab:running_times_inpainting}
    \vspace{-3mm}
    \small
    \begin{tabular}{@{}C{2.2cm}@{}| 
                    @{}C{2.4cm}@{} @{}C{1.7cm}@{} @{}C{1.7cm}@{} @{}C{2.2cm}@{}
                    @{}C{2.0cm}@{} @{}C{2.5cm}@{} @{}C{1.8cm}@{} @{}C{1.6cm}@{}}
        \Xhline{0.8pt}
        Methods     & DeepFillv2~\cite{yu2019free}  &DSI~\cite{peng2021generating}   & LaMa~\cite{suvorov2022resolution}   & RePaint~\cite{Lugmayr_2022_CVPR}   & DDRM~\cite{kawar2022denoising}   & Score-SDE~\cite{song2020score}   & MCG~\cite{chung2022improving}   & \textit{DifFace} \\
        \Xhline{0.4pt}
        Runtime (s)   & 0.05        & 29.32  & 0.02   & 139.20    & 11.74  & 11.67   & 28.09  & 3.58 \\
        \Xhline{0.8pt}
    \end{tabular}
\end{table*}
\begin{table*}[t]
    \centering
    \caption{Comparisons of different methods with respect to running time (s) on the task of BFR. Note that the running time is counted on the NVIDIA Tesla V100 with a resolution of $512\times 512$.}
    \label{tab:running_times_bfr}
    \vspace{-3mm}
    \small
    \begin{tabular}{@{}C{1.7cm}@{}| 
                    @{}C{1.8cm}@{} @{}C{1.9cm}@{} @{}C{2.3cm}@{} @{}C{2.1cm}@{}
                    @{}C{2.6cm}@{} @{}C{1.8cm}@{} @{}C{2.2cm}@{} @{}C{1.6cm}@{}}
        \Xhline{0.8pt}
        Methods      & DFDNet~\cite{li2020blind} & PULSE~\cite{menon2020pulse}  & PSFRGAN~\cite{chen2021progressive}   & GFPGAN~\cite{wang2021towards}   & RestoreFormer~\cite{wang2022restoreformer}   & VQFR~\cite{gu2022vqfr}   & CodeFormer~\cite{toward2022zhou}   & \textit{DifFace} \\
        \Xhline{0.4pt}
        Runtime (s)   & 1.40   & 3.93   & 0.06      & 0.04    & 0.14     & 0.24   & 0.08  & 4.32 \\
        \Xhline{0.8pt}
    \end{tabular}
\end{table*}
\begin{table*}[!t]
    \centering
    \caption{Performance comparisons of DFDNet, CodeFormer, and \textit{DifFace} under different acceleration settings. ``\textit{Dif}(A/B)'' means that the whole reverse process of the pre-trained diffusion model contains ``A'' sampling steps after acceleration, and the starting timestep $N$ in \textit{DifFace} is set as ``B''. In the main text of this paper, we reported the performance of \textit{Dif}(250/100) as our result. It should be noted that the employed diffusion model was maintained at a length of 1000 during the training phase.}
    \label{tab:acceleration}
    \vspace{-3mm}
    \small
    \begin{tabular}{@{}C{2.6cm}@{}|
                    @{}C{2.4cm}@{} @{}C{2.2cm}@{} @{}C{2.1cm}@{} @{}C{2.0cm}@{} @{}C{2.0cm}@{} ||
                    @{}C{2.0cm}@{} @{}C{2.6cm}@{} }
        \Xhline{0.8pt}
        \multirow{2}*{Metrics} & \multicolumn{7}{c}{Methods} \\
        \Xcline{2-8}{0.4pt}
                            & \textit{Dif}(500/200)   & \textit{Dif}(250/100) & \textit{Dif}(100/40) & \textit{Dif}(50/20)
                            & \textit{Dif}(20/8)   & DFDNet~\cite{li2020blind}    & CodeFormer~\cite{toward2022zhou} \\
        \Xhline{0.4pt}
        PSNR$\uparrow$      & 23.43    & 23.44   & 23.41   & 23.21    & 21.88   & 22.39  & 22.70 \\
        SSIM$\uparrow$      & 0.690    & 0.690   & 0.684   & 0.657    & 0.496   & 0.627  & 0.644 \\
        LPIPS$\downarrow$   & 0.459    & 0.461   & 0.475   & 0.498    & 0.591   & 0.582  & 0.438 \\
        IDS$\downarrow$     & 65.02    & 64.94   & 64.87   & 64.91    & 65.86   & 86.65  & 64.64 \\
        LMD$\downarrow$     & 6.06     & 6.06    & 6.07    & 6.06     & 6.10    & 22.42  & 8.01 \\
        FID-F$\downarrow$   & 49.22    & 48.98   & 49.72   & 54.32    & 70.89   & 96.90  & 61.31 \\
        FID-G$\downarrow$   & 20.28    & 20.29   & 21.74   & 26.35    & 42.54   & 85.14  & 25.99 \\
        \hline \hline                                                                           
        Runntime (s)    & 8.58     & 4.32    & 1.77    & 0.92     & 0.41    & 1.40   & 0.08 \\
        \Xhline{0.8pt}
    \end{tabular}
\end{table*}

Furthermore, we provide a series of visual illustrations encompassing various mask types in Fig.~\ref{fig:syn_inpainting}. \zsyrevise{In the case of ``Box" mask}, all methods perform well with the exception of DeepFillv2~\cite{yu2019free}. When the occluded area in the mask increase, GAN-based methods tend to generate obvious artifacts, \zsyrevise{particularly when confronted with the ``Expand" mask}. In contrast, the diffusion-based approaches, such as RePaint~\cite{Lugmayr_2022_CVPR}, MCG~\cite{chung2022improving}, and our proposed \textit{DifFace}, consistently yield more plausible and realistic results under this scenario. Notably,  \textit{DifFace} distinguishes itself by excelling in the preservation of coherency to the unmasked regions, as demonstrated in the last three examples. This superiority can be largely attributed to the elaborate optimization technique in Sec.~\ref{subsec:exp_hyper_inpainting}. The qualitative analysis presented herein reaffirms the stability and exceptional performance of \textit{DifFace}, aligning with the quantitative comparison above. 
\begin{figure}[t]
    \centering
    \includegraphics[width=\linewidth]{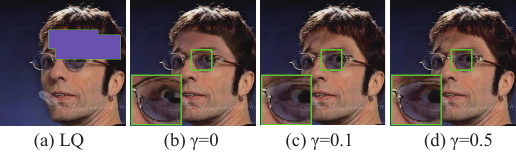}
    \vspace{-6mm}
    \caption{Recovered results of \textit{DifFace} under different settings on the hyper-parameter $\gamma$ of Eq.~\eqref{eq:inparing_optimization}.}
    \label{fig:gamma-inpainting}
\end{figure}

\subsection{Hyper-parameter Analysis} \label{subsec:exp_hyper_inpainting}
For the task of inpainting, we have devised an associated optimization strategy to further refine the recovered results during sampling as introduced in Sec.~\ref{sec:extension}. The specific optimization problem is formally articulated in Eq.~\eqref{eq:inparing_optimization}, involving in a key hyper-parameter $\gamma$. This hyper-parameter controls the degree of influence exerted by the priori knowledge of the degradation model. To exemplify the impact of this hyper-parameter, we present the visual results obtained by our method under various $\gamma$ configurations in Figure~\ref{fig:gamma-inpainting}. When $\gamma$ is set to zero, thus omitting the optimized procedure, discernible incoherence emerges between the recovered regions and the unmasked areas. However, this inconsistency can be notably mitigated by increasing the value of $\gamma$. This observation indicates the importance and effectiveness of the proposed strategy in addressing the incoherence issue.

\section{Limitation}
Despite the commendable performance of \textit{DifFace}, the inference efficiency of our method is limited by the iterative sampling process inherited from the diffusion model. To provide a quantitative evaluation, we present a comparison on the running time of \textit{DifFace} alongside other competitive methods in Table~\ref{tab:running_times_inpainting} and Table~\ref{tab:running_times_bfr}, which are conducted on the tasks of face inpainting and face restoration, respectively. 

Upon reviewing the results of~Table~\ref{tab:running_times_inpainting} with regards to inpainting, it is evident that \textit{DifFace} exhibits conspicuous advantages in terms of efficiency among the diffusion-based approaches, inluding RePaint~\cite{Lugmayr_2022_CVPR}, DDRM~\cite{kawar2022denoising}, Score-SDE~\cite{song2020score}, and MCG~\cite{chung2022improving}. Unfortunately, it still maintains a comparatively slower runtime when compared to CNN-based methodologies of DeepFillv2~\cite{yu2018bisenet} and LaMa~\cite{suvorov2022resolution}. Nonetheless, in light of its consistently superior performance, it is well-founded that our proposed \textit{DifFace} holds great and practical promise for inpainting. 

For the task of BFR, the corresponding efficiency assessment is summarized in Table~\ref{tab:running_times_bfr}. \textit{DifFace} demonstrates competitiveness with the optimization-based approach PULSE~\cite{menon2020pulse} in terms of speed, while exhibiting slower performance compared to other GAN-based methodologies. It is worth highlighting that \textit{DifFace} can be further accelerated through DDIM~\cite{song2021denoising}. Specifically, Table~\ref{tab:acceleration} offers a comprehensive performance analysis of \textit{DifFace} on the CelebA-Test dataset under various acceleration settings. We can see that it is possible to reduce the sampling process to 20 steps without evident performance drop. Under this setting, \textit{DifFace} is still comparable to CodeFormer~\cite{toward2022zhou}. This cuts the running time to 0.92s, falling between DFDNet~\cite{li2020blind} and CodeFormer~\cite{toward2022zhou}. We are committed to exploring more advanced acceleration techniques in our future research endeavors to further enhance inference speed.


\section{Conclusion}
We have proposed a new BFR method called \textit{DifFace} in this work. \textit{DifFace} is appealing as it only relies on a restoration backbone that is trained with $L_1$ loss. This vastly simplifies the complicated training objectives in most current approaches. Importantly, we have proposed a posterior distribution that is well-suited for BFR. It consists of a transition kernel and a Markov chain partially borrowed from a pre-trained diffusion model. The former acts as an error compressor, and thus makes our method more robust to severe degradations. The latter effectively leverages the powerful diffusion model to facilitate BFR. Extensive experiments on the face restoration and face inpainting have demonstrated the effectiveness and robustness of our method both on synthetic and real-world datasets. We hope that this work could inspire more robust diffusion-based restoration methods in the future.

 
\bibliography{reference}

\begin{thebibliography}{10}
\providecommand{\url}[1]{#1}
\csname url@samestyle\endcsname
\providecommand{\newblock}{\relax}
\providecommand{\bibinfo}[2]{#2}
\providecommand{\BIBentrySTDinterwordspacing}{\spaceskip=0pt\relax}
\providecommand{\BIBentryALTinterwordstretchfactor}{4}
\providecommand{\BIBentryALTinterwordspacing}{\spaceskip=\fontdimen2\font plus
\BIBentryALTinterwordstretchfactor\fontdimen3\font minus
  \fontdimen4\font\relax}
\providecommand{\BIBforeignlanguage}[2]{{%
\expandafter\ifx\csname l@#1\endcsname\relax
\typeout{** WARNING: IEEEtran.bst: No hyphenation pattern has been}%
\typeout{** loaded for the language `#1'. Using the pattern for}%
\typeout{** the default language instead.}%
\else
\language=\csname l@#1\endcsname
\fi
#2}}
\providecommand{\BIBdecl}{\relax}
\BIBdecl

\bibitem{wang2021towards}
X.~Wang, Y.~Li, H.~Zhang, and Y.~Shan, ``Towards real-world blind face
  restoration with generative facial prior,'' in \emph{Proceedings of the
  IEEE/CVF Conference on Computer Vision and Pattern Recognition (CVPR)}, 2021,
  pp. 9168--9178.

\bibitem{tu2021joint}
X.~Tu, J.~Zhao, Q.~Liu, W.~Ai, G.~Guo, Z.~Li, W.~Liu, and J.~Feng, ``Joint face
  image restoration and frontalization for recognition,'' \emph{IEEE
  Transactions on Circuits and Systems for Video Technology (TCSVT)}, vol.~32,
  no.~3, pp. 1285--1298, 2021.

\bibitem{feihong2022toward}
L.~Feihong, C.~Hang, L.~Kang, D.~Qiliang, Z.~jian, Z.~Kaipeng, and H.~Hong*,
  ``Toward high-quality face-mask occluded restoration,'' \emph{ACM
  Transactions on Multimedia Computing, Communications, and Applications
  (TOMM)}, 2022.

\bibitem{gu2022vqfr}
Y.~Gu, X.~Wang, L.~Xie, C.~Dong, G.~Li, Y.~Shan, and M.-M. Cheng, ``{VQFR:}
  blind face restoration with vector-quantized dictionary and parallel
  decoder,'' in \emph{Proceedings of the European Conference on Computer Vision
  (ECCV)}, 2022.

\bibitem{goodfellow2014generative}
I.~Goodfellow, J.~Pouget-Abadie, M.~Mirza, B.~Xu, D.~Warde-Farley, S.~Ozair,
  A.~Courville, and Y.~Bengio, ``Generative adversarial nets,'' in
  \emph{Proceedings of Advances in Neural Information Processing Systems
  (NeurIPS)}, 2014.

\bibitem{johnson2016perceptual}
J.~Johnson, A.~Alahi, and L.~Fei-Fei, ``Perceptual losses for real-time style
  transfer and super-resolution,'' in \emph{Proceedings of the European
  Conference on Computer Vision (ECCV)}, 2016, pp. 694--711.

\bibitem{zhang2018unreasonable}
R.~Zhang, P.~Isola, A.~A. Efros, E.~Shechtman, and O.~Wang, ``The unreasonable
  effectiveness of deep features as a perceptual metric,'' in \emph{Proceedings
  of the IEEE/CVF Conference on Computer Vision and Pattern Recognition
  (CVPR)}, 2018, pp. 586--595.

\bibitem{chen2018fsrnet}
Y.~Chen, Y.~Tai, X.~Liu, C.~Shen, and J.~Yang, ``{FSRNet}: End-to-end learning
  face super-resolution with facial priors,'' in \emph{Proceedings of the
  IEEE/CVF Conference on Computer Vision and Pattern Recognition (CVPR)}, 2018,
  pp. 2492--2501.

\bibitem{li2020blind}
X.~Li, C.~Chen, S.~Zhou, X.~Lin, W.~Zuo, and L.~Zhang, ``Blind face restoration
  via deep multi-scale component dictionaries,'' in \emph{Proceedings of the
  European Conference on Computer Vision (ECCV)}, 2020, pp. 399--415.

\bibitem{chan2021glean}
K.~C. Chan, X.~Wang, X.~Xu, J.~Gu, and C.~C. Loy, ``{GLEAN:} generative latent
  bank for large-factor image super-resolution,'' \emph{IEEE Transactions on
  Pattern Analysis and Machine Intelligence}, 2022.

\bibitem{pan2021exploiting}
X.~Pan, X.~Zhan, B.~Dai, D.~Lin, C.~C. Loy, and P.~Luo, ``Exploiting deep
  generative prior for versatile image restoration and manipulation,''
  \emph{IEEE Transactions on Pattern Analysis and Machine Intelligence}, 2021.

\bibitem{yang2021gan}
T.~Yang, P.~Ren, X.~Xie, and L.~Zhang, ``{GAN} prior embedded network for blind
  face restoration in the wild,'' in \emph{Proceedings of the IEEE/CVF
  Conference on Computer Vision and Pattern Recognition (CVPR)}, 2021, pp.
  672--681.

\bibitem{toward2022zhou}
S.~Zhou, K.~C.~K. Chan, C.~Li, and C.~C. Loy, ``Towards robust blind face
  restoration with codebook lookup transformer,'' \emph{Proceedings of Advances
  in Neural Information Processing Systems (NeurIPS)}, 2022.

\bibitem{sohl2015deep}
J.~Sohl-Dickstein, E.~Weiss, N.~Maheswaranathan, and S.~Ganguli, ``Deep
  unsupervised learning using nonequilibrium thermodynamics,'' in
  \emph{International Conference on Machine Learning (ICML)}, 2015, pp.
  2256--2265.

\bibitem{dhariwal2021diffusion}
P.~Dhariwal and A.~Nichol, ``Diffusion models beat {GANs} on image synthesis,''
  in \emph{Proceedings of Advances in Neural Information Processing Systems
  (NeurIPS)}, 2021, pp. 8780--8794.

\bibitem{ho2020denoising}
J.~Ho, A.~Jain, and P.~Abbeel, ``Denoising diffusion probabilistic models,''
  \emph{Advances in Neural Information Processing Systems (NeurIPS)}, vol.~33,
  pp. 6840--6851, 2020.

\bibitem{saharia2022image}
C.~Saharia, J.~Ho, W.~Chan, T.~Salimans, D.~J. Fleet, and M.~Norouzi, ``Image
  super-resolution via iterative refinement,'' \emph{IEEE Transactions on
  Pattern Analysis and Machine Intelligence}, 2022.

\bibitem{chen2021progressive}
C.~Chen, X.~Li, L.~Yang, X.~Lin, L.~Zhang, and K.-Y.~K. Wong, ``Progressive
  semantic-aware style transformation for blind face restoration,'' in
  \emph{Proceedings of the IEEE/CVF Conference on Computer Vision and Pattern
  Recognition (CVPR)}, 2021, pp. 11\,896--11\,905.

\bibitem{shen2018deep}
Z.~Shen, W.-S. Lai, T.~Xu, J.~Kautz, and M.-H. Yang, ``Deep semantic face
  deblurring,'' in \emph{Proceedings of the IEEE/CVF Conference on Computer
  Vision and Pattern Recognition (CVPR)}, 2018, pp. 8260--8269.

\bibitem{ren2019face}
W.~Ren, J.~Yang, S.~Deng, D.~Wipf, X.~Cao, and X.~Tong, ``Face video deblurring
  using {3D} facial priors,'' in \emph{Proceedings of the IEEE/CVF
  International Conference on Computer Vision (ICCV)}, 2019, pp. 9388--9397.

\bibitem{hu2020face}
X.~Hu, W.~Ren, J.~LaMaster, X.~Cao, X.~Li, Z.~Li, B.~Menze, and W.~Liu, ``Face
  super-resolution guided by {3D} facial priors,'' in \emph{Proceedings of the
  European Conference on Computer Vision (ECCV)}, 2020, pp. 763--780.

\bibitem{zhu2022blind}
F.~Zhu, J.~Zhu, W.~Chu, X.~Zhang, X.~Ji, C.~Wang, and Y.~Tai, ``Blind face
  restoration via integrating face shape and generative priors,'' in
  \emph{Proceedings of the IEEE/CVF Conference on Computer Vision and Pattern
  Recognition (CVPR)}, 2022, pp. 7662--7671.

\bibitem{li2018learning}
X.~Li, M.~Liu, Y.~Ye, W.~Zuo, L.~Lin, and R.~Yang, ``Learning warped guidance
  for blind face restoration,'' in \emph{Proceedings of the European Conference
  on Computer Vision (ECCV)}, 2018, pp. 272--289.

\bibitem{dogan2019exemplar}
B.~Dogan, S.~Gu, and R.~Timofte, ``Exemplar guided face image super-resolution
  without facial landmarks,'' in \emph{Proceedings of the IEEE/CVF
  International Conference on Computer Vision Workshops (CVPR-W)}, 2019, pp.
  0--0.

\bibitem{xia2022gan}
W.~Xia, Y.~Zhang, Y.~Yang, J.-H. Xue, B.~Zhou, and M.-H. Yang, ``{GAN}
  inversion: A survey,'' \emph{IEEE Transactions on Pattern Analysis and
  Machine Intelligence}, 2022.

\bibitem{menon2020pulse}
S.~Menon, A.~Damian, S.~Hu, N.~Ravi, and C.~Rudin, ``{PULSE:} self-supervised
  photo upsampling via latent space exploration of generative models,'' in
  \emph{Proceedings of the IEEE/CVF Conference on Computer Vision and Pattern
  Recognition (CVPR)}, 2020, pp. 2437--2445.

\bibitem{gu2020image}
J.~Gu, Y.~Shen, and B.~Zhou, ``Image processing using multi-code {GAN} prior,''
  in \emph{Proceedings of the IEEE/CVF Conference on Computer Vision and
  Pattern Recognition (CVPR)}, 2020, pp. 3012--3021.

\bibitem{mei2023ltt}
K.~Mei and V.~M. Patel, ``{LTT-GAN:}: Looking through turbulence by inverting
  gans,'' \emph{IEEE Journal of Selected Topics in Signal Processing}, 2023.

\bibitem{karras2019style}
T.~Karras, S.~Laine, and T.~Aila, ``A style-based generator architecture for
  generative adversarial networks,'' in \emph{Proceedings of the IEEE/CVF
  Conference on Computer Vision and Pattern Recognition (CVPR)}, 2019, pp.
  4401--4410.

\bibitem{esser2021taming}
P.~Esser, R.~Rombach, and B.~Ommer, ``Taming transformers for high-resolution
  image synthesis,'' in \emph{Proceedings of the IEEE/CVF Conference on
  Computer Vision and Pattern Recognition (CVPR)}, 2021, pp. 12\,873--12\,883.

\bibitem{wang2022restoreformer}
Z.~Wang, J.~Zhang, R.~Chen, W.~Wang, and P.~Luo, ``{RestoreFormer:}
  high-quality blind face restoration from undegraded key-value pairs,'' in
  \emph{Proceedings of the IEEE/CVF Conference on Computer Vision and Pattern
  Recognition (CVPR)}, 2022, pp. 17\,512--17\,521.

\bibitem{li2022srdiff}
H.~Li, Y.~Yang, M.~Chang, S.~Chen, H.~Feng, Z.~Xu, Q.~Li, and Y.~Chen,
  ``{SRDiff:} single image super-resolution with diffusion probabilistic
  models,'' \emph{Neurocomputing}, vol. 479, pp. 47--59, 2022.

\bibitem{zhao2023towards}
Y.~Zhao, T.~Hou, Y.-C. Su, X.~Jia, Y.~Li, and M.~Grundmann, ``Towards authentic
  face restoration with iterative diffusion models and beyond,'' in
  \emph{Proceedings of the IEEE/CVF International Conference on Computer Vision
  (ICCV)}, 2023, pp. 7312--7322.

\bibitem{rombach2022high}
R.~Rombach, A.~Blattmann, D.~Lorenz, P.~Esser, and B.~Ommer, ``High-resolution
  image synthesis with latent diffusion models,'' in \emph{Proceedings of the
  IEEE/CVF Conference on Computer Vision and Pattern Recognition (CVPR)}, 2022,
  pp. 10\,684--10\,695.

\bibitem{Choi_2021_ICCV}
J.~Choi, S.~Kim, Y.~Jeong, Y.~Gwon, and S.~Yoon, ``Ilvr: Conditioning method
  for denoising diffusion probabilistic models,'' in \emph{Proceedings of the
  IEEE/CVF International Conference on Computer Vision (ICCV)}, October 2021,
  pp. 14\,367--14\,376.

\bibitem{Lugmayr_2022_CVPR}
A.~Lugmayr, M.~Danelljan, A.~Romero, F.~Yu, R.~Timofte, and L.~Van~Gool,
  ``Repaint: Inpainting using denoising diffusion probabilistic models,'' in
  \emph{Proceedings of the IEEE/CVF Conference on Computer Vision and Pattern
  Recognition (CVPR)}, June 2022, pp. 11\,461--11\,471.

\bibitem{kawar2022denoising}
B.~Kawar, M.~Elad, S.~Ermon, and J.~Song, ``Denoising diffusion restoration
  models,'' in \emph{Proceedings of Advances in Neural Information Processing
  Systems (NeurIPS)}, vol.~35, 2022, pp. 23\,593--23\,606.

\bibitem{chung2022come}
H.~Chung, B.~Sim, and J.~C. Ye, ``Come-closer-diffuse-faster: Accelerating
  conditional diffusion models for inverse problems through stochastic
  contraction,'' in \emph{Proceedings of the IEEE/CVF Conference on Computer
  Vision and Pattern Recognition (CVPR)}, 2022, pp. 12\,413--12\,422.

\bibitem{chung2022improving}
H.~Chung, B.~Sim, D.~Ryu, and J.~C. Ye, ``Improving diffusion models for
  inverse problems using manifold constraints,'' in \emph{Proceedings of
  Advances in Neural Information Processing Systems (NeurIPS)}, vol.~35, 2022,
  pp. 25\,683--25\,696.

\bibitem{Fei_2023_CVPR}
B.~Fei, Z.~Lyu, L.~Pan, J.~Zhang, W.~Yang, T.~Luo, B.~Zhang, and B.~Dai,
  ``Generative diffusion prior for unified image restoration and enhancement,''
  in \emph{Proceedings of the IEEE/CVF Conference on Computer Vision and
  Pattern Recognition (CVPR)}, June 2023, pp. 9935--9946.

\bibitem{wang2023dr2}
Z.~Wang, Z.~Zhang, X.~Zhang, H.~Zheng, M.~Zhou, Y.~Zhang, and Y.~Wang, ``Dr2:
  Diffusion-based robust degradation remover for blind face restoration,'' in
  \emph{Proceedings of the IEEE/CVF Conference on Computer Vision and Pattern
  Recognition (CVPR)}, 2023, pp. 1704--1713.

\bibitem{wang2023zeroshot}
Y.~Wang, J.~Yu, and J.~Zhang, ``Zero-shot image restoration using denoising
  diffusion null-space model,'' in \emph{Proceedings of International
  Conference on Learning Representations (ICLR)}, 2023.

\bibitem{zhu2023denoising}
Y.~Zhu, K.~Zhang, J.~Liang, J.~Cao, B.~Wen, R.~Timofte, and L.~Van~Gool,
  ``Denoising diffusion models for plug-and-play image restoration,'' in
  \emph{Proceedings of the IEEE/CVF International Conference on Computer Vision
  Workshops (CVPR-W)}, 2023, pp. 1219--1229.

\bibitem{yue2023resshift}
Z.~Yue, J.~Wang, and C.~C. Loy, ``Resshift: Efficient diffusion model for image
  super-resolution by residual shifting,'' in \emph{Proceedings of Advances in
  Neural Information Processing Systems (NeurIPS)}, 2023.

\bibitem{wang2023exploiting}
J.~Wang, Z.~Yue, S.~Zhou, K.~C. Chan, and C.~C. Loy, ``Exploiting diffusion
  prior for real-world image super-resolution,'' \emph{arXiv preprint
  arXiv:2305.07015}, 2023.

\bibitem{lin2023diffbir}
X.~Lin, J.~He, Z.~Chen, Z.~Lyu, B.~Fei, B.~Dai, W.~Ouyang, Y.~Qiao, and
  C.~Dong, ``Diffbir: Towards blind image restoration with generative diffusion
  prior,'' \emph{arXiv preprint arXiv:2308.15070}, 2023.

\bibitem{qiu2023diffbfr}
X.~Qiu, C.~Han, Z.~Zhang, B.~Li, T.~Guo, and X.~Nie, ``{DiffBFR}: Bootstrapping
  diffusion model for blind face restoration,'' in \emph{Proceedings of the ACM
  International Conference on Multimedia (ACM MM)}, 2023, pp. 7785--7795.

\bibitem{song2021denoising}
J.~Song, C.~Meng, and S.~Ermon, ``Denoising diffusion implicit models,'' in
  \emph{Proceedings of International Conference on Learning Representations
  (ICLR)}, 2021.

\bibitem{bishop2006pattern}
C.~M. Bishop and N.~M. Nasrabadi, \emph{Pattern recognition and machine
  learning}.\hskip 1em plus 0.5em minus 0.4em\relax Springer, 2006, vol.~4,
  no.~4.

\bibitem{dong2015image}
C.~Dong, C.~C. Loy, K.~He, and X.~Tang, ``Image super-resolution using deep
  convolutional networks,'' \emph{IEEE Transactions on Pattern Analysis and
  Machine Intelligence}, vol.~38, no.~2, pp. 295--307, 2015.

\bibitem{liang2021swinir}
J.~Liang, J.~Cao, G.~Sun, K.~Zhang, L.~Van~Gool, and R.~Timofte, ``{SwinIR:}
  image restoration using swin transformer,'' in \emph{Proceedings of the
  IEEE/CVF International Conference on Computer Vision Workshops (ICCV-W)},
  2021, pp. 1833--1844.

\bibitem{kingma2015adam}
D.~P. Kingma and J.~Ba, ``Adam: A method for stochastic optimization,'' in
  \emph{Proceedings of International Conference on Learning Representations
  (ICLR)}, 2015.

\bibitem{paszke2019pytorch}
A.~Paszke, S.~Gross, F.~Massa, A.~Lerer, J.~Bradbury, G.~Chanan, T.~Killeen,
  Z.~Lin, N.~Gimelshein, L.~Antiga \emph{et~al.}, ``{PyTorch:} an imperative
  style, high-performance deep learning library,'' in \emph{Advances in Neural
  Information Processing Systems (NeurIPS)}, vol.~32, 2019.

\bibitem{DBLP:conf/iclr/LoshchilovH17}
I.~Loshchilov and F.~Hutter, ``{SGDR:} stochastic gradient descent with warm
  restarts,'' in \emph{Proceedings of International Conference on Learning
  Representations (ICLR)}, 2017.

\bibitem{karras2018progressive}
T.~Karras, T.~Aila, S.~Laine, and J.~Lehtinen, ``Progressive growing of {GANs}
  for improved quality, stability, and variation,'' in \emph{Proceedings of the
  International Conference on Learning Representations (ICLR)}, 2018.

\bibitem{nichol2021improved}
A.~Q. Nichol and P.~Dhariwal, ``Improved denoising diffusion probabilistic
  models,'' in \emph{International Conference on Machine Learning
  (ICML)}.\hskip 1em plus 0.5em minus 0.4em\relax PMLR, 2021, pp. 8162--8171.

\bibitem{huang2008labeled}
G.~B. Huang, M.~Mattar, T.~Berg, and E.~Learned-Miller, ``Labeled faces in the
  wild: A database forstudying face recognition in unconstrained
  environments,'' in \emph{Workshop on faces in 'Real-Life' Images: detection,
  alignment, and recognition}, 2008.

\bibitem{yang2016wider}
S.~Yang, P.~Luo, C.-C. Loy, and X.~Tang, ``Wider face: A face detection
  benchmark,'' in \emph{Proceedings of the IEEE/CVF Conference on Computer
  Vision and Pattern Recognition (CVPR)}, 2016, pp. 5525--5533.

\bibitem{wang2004image}
Z.~Wang, A.~C. Bovik, H.~R. Sheikh, and E.~P. Simoncelli, ``Image quality
  assessment: from error visibility to structural similarity,'' \emph{IEEE
  Transactions on Image Processing}, vol.~13, no.~4, pp. 600--612, 2004.

\bibitem{heusel2017gans}
M.~Heusel, H.~Ramsauer, T.~Unterthiner, B.~Nessler, and S.~Hochreiter, ``{GANs}
  trained by a two time-scale update rule converge to a local nash
  equilibrium,'' \emph{Advances in Neural Information Processing Systems
  (NeurIPS}, vol.~30, 2017.

\bibitem{deng2019arcface}
J.~Deng, J.~Guo, N.~Xue, and S.~Zafeiriou, ``{ArcFace:} additive angular margin
  loss for deep face recognition,'' in \emph{Proceedings of the IEEE/CVF
  Conference on Computer Vision and Pattern Recognition (CVPR)}, 2019, pp.
  4690--4699.

\bibitem{mittal2012making}
A.~Mittal, R.~Soundararajan, and A.~C. Bovik, ``Making a “completely blind”
  image quality analyzer,'' \emph{IEEE Signal Processing Letters}, vol.~20,
  no.~3, pp. 209--212, 2012.

\bibitem{ma2017learning}
C.~Ma, C.-Y. Yang, X.~Yang, and M.-H. Yang, ``Learning a no-reference quality
  metric for single-image super-resolution,'' \emph{Computer Vision and Image
  Understanding}, vol. 158, pp. 1--16, 2017.

\bibitem{blau20182018}
Y.~Blau, R.~Mechrez, R.~Timofte, T.~Michaeli, and L.~Zelnik-Manor, ``The 2018
  {PIRM} challenge on perceptual image super-resolution,'' in \emph{Proceedings
  of the European Conference on Computer Vision Workshops (ECCV-W)}, 2018, pp.
  0--0.

\bibitem{yu2019free}
J.~Yu, Z.~Lin, J.~Yang, X.~Shen, X.~Lu, and T.~S. Huang, ``Free-form image
  inpainting with gated convolution,'' in \emph{Proceedings of the IEEE/CVF
  International Conference on Computer Vision (ICCV)}, 2019, pp. 4471--4480.

\bibitem{peng2021generating}
J.~Peng, D.~Liu, S.~Xu, and H.~Li, ``Generating diverse structure for image
  inpainting with hierarchical vq-vae,'' in \emph{Proceedings of the IEEE/CVF
  Conference on Computer Vision and Pattern Recognition (CVPR)}, 2021, pp.
  10\,775--10\,784.

\bibitem{suvorov2022resolution}
R.~Suvorov, E.~Logacheva, A.~Mashikhin, A.~Remizova, A.~Ashukha, A.~Silvestrov,
  N.~Kong, H.~Goka, K.~Park, and V.~Lempitsky, ``Resolution-robust large mask
  inpainting with fourier convolutions,'' in \emph{Proceedings of the IEEE/CVF
  Winter Conference on Applications of Computer Vision (WACV)}, 2022, pp.
  2149--2159.

\bibitem{song2020score}
Y.~Song, J.~Sohl-Dickstein, D.~P. Kingma, A.~Kumar, S.~Ermon, and B.~Poole,
  ``Score-based generative modeling through stochastic differential
  equations,'' in \emph{Proceedings of the International Conference on Learning
  Representations (ICLR)}, 2020.

\bibitem{yu2018bisenet}
C.~Yu, J.~Wang, C.~Peng, C.~Gao, G.~Yu, and N.~Sang, ``Bisenet: Bilateral
  segmentation network for real-time semantic segmentation,'' in
  \emph{Proceedings of the European Conference on Computer Vision (ECCV)},
  2018, pp. 325--341.

\bibitem{shi2016real}
W.~Shi, J.~Caballero, F.~Husz{\'a}r, J.~Totz, A.~P. Aitken, R.~Bishop,
  D.~Rueckert, and Z.~Wang, ``Real-time single image and video super-resolution
  using an efficient sub-pixel convolutional neural network,'' in
  \emph{Proceedings of the IEEE Conference on Computer Vision and Pattern
  Recognition (CVPR)}, 2016, pp. 1874--1883.

\bibitem{he2016deep}
K.~He, X.~Zhang, S.~Ren, and J.~Sun, ``Deep residual learning for image
  recognition,'' in \emph{Proceedings of the IEEE/CVF Conference on Computer
  Vision and Pattern Recognition (CVPR)}, 2016, pp. 770--778.

\bibitem{song2022pseudoinverse}
J.~Song, A.~Vahdat, M.~Mardani, and J.~Kautz, ``Pseudoinverse-guided diffusion
  models for inverse problems,'' in \emph{Proceedings of International
  Conference on Learning Representations (ICLR)}, 2022.

\bibitem{deng2009imagenet}
J.~Deng, W.~Dong, R.~Socher, L.-J. Li, K.~Li, and L.~Fei-Fei, ``{ImageNet:} a
  large-scale hierarchical image database,'' in \emph{Proceedings of the
  IEEE/CVF Conference on Computer Vision and Pattern Recognition (CVPR)}.\hskip
  1em plus 0.5em minus 0.4em\relax Ieee, 2009, pp. 248--255.

\end{thebibliography}
\bibliographystyle{IEEEtran}



\vspace{-10cm}
\begin{IEEEbiography}[{\includegraphics[width=1in,height=1.25in,clip,keepaspectratio]{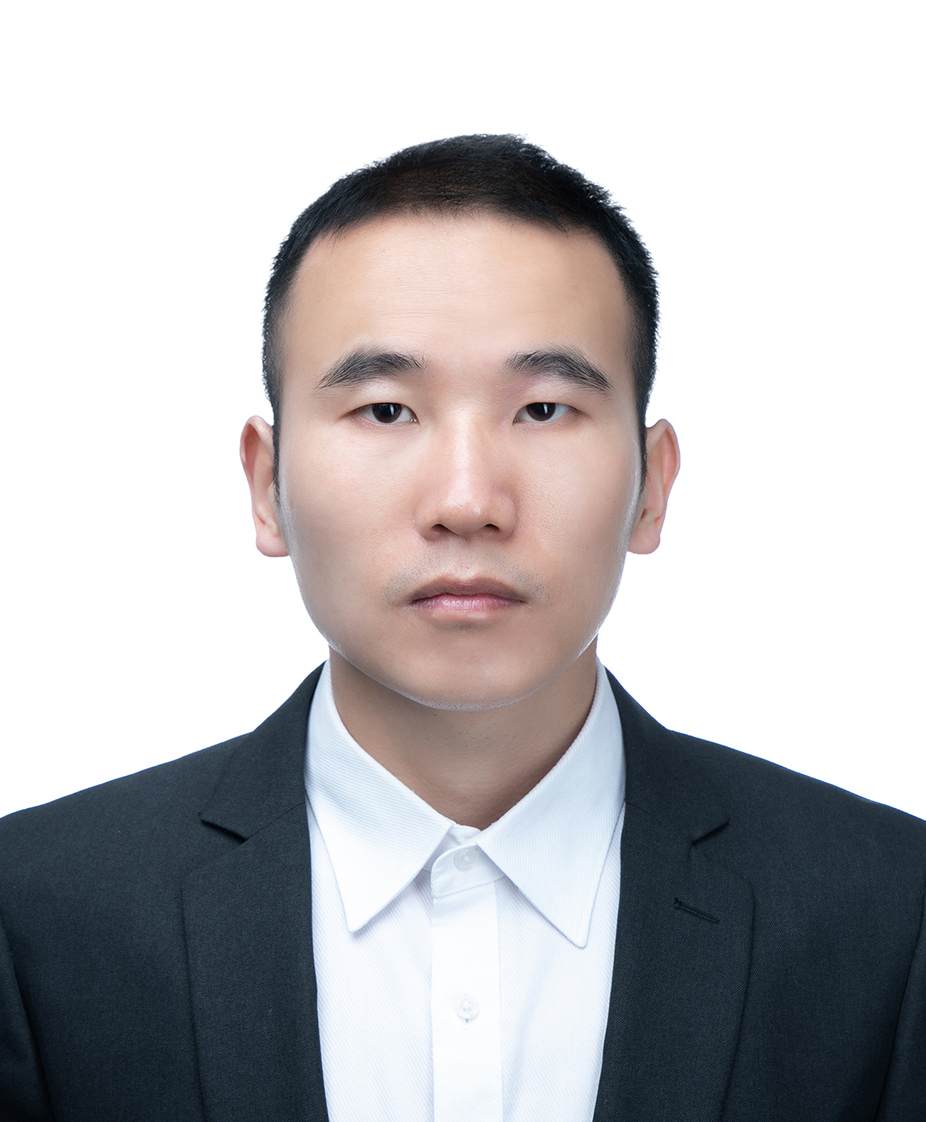}}]{Zongsheng Yue}
(Member, IEEE) received his Ph.D. degree from Xi'an Jiaotong University, Xi'an, China, in 2021. He is currently a postdoctoral research fellow with school of Computer Science and Engineering, Nanyang Technological University. From September 2021 to March 2022, he was a associate research in the Department of Computer Science, Hong Kong University. He was a research assistant in the Department of Computing, Hong Kong Polytechnic University during October 2018 to June 2019 and the Institute of Future Cities, The Chinese University of Hong Kong during February 2017 to September 2017, respectively. His current research interests include noise modeling, image restoration, and diffusion model.
\end{IEEEbiography}
 
\vspace{-11cm}

\begin{IEEEbiography}[{\includegraphics[width=1in,height=1.25in,clip,keepaspectratio]{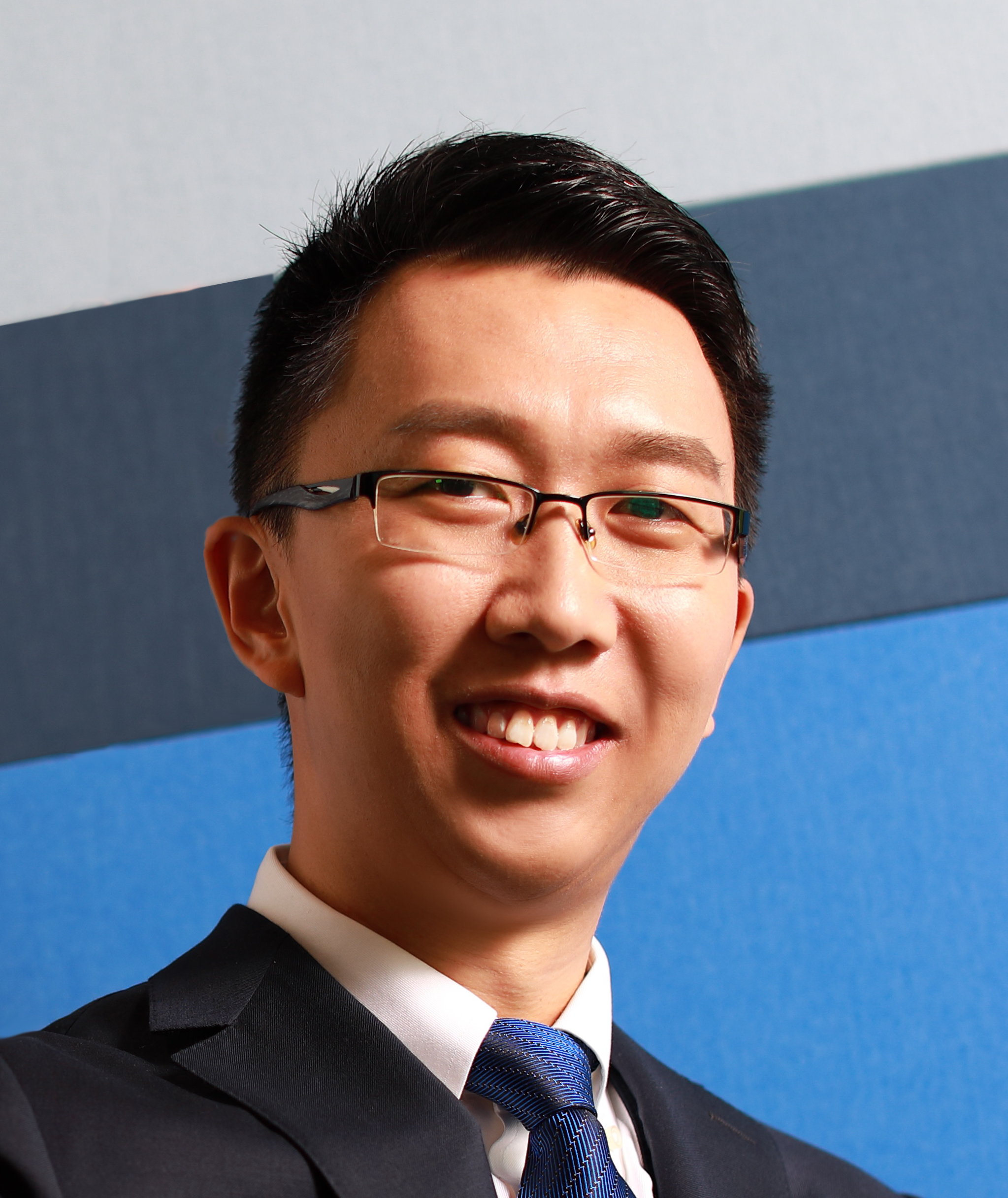}}]{Chen Change Loy}
(Senior Member, IEEE) is currently a Professor with the School of Computer Science and Engineering, Nanyang Technological University, Singapore. He received the PhD degree in computer science from the Queen Mary University of London, in 2010. Prior to joining NTU, he served as a research assistant professor with the Department of Information Engineering, The Chinese University of Hong Kong, from 2013 to 2018. His research interests include computer vision and deep learning with a focus on image/video restoration and enhancement, generative tasks, and representation learning. He serves as an associate editor of the IEEE Transactions on Pattern Analysis and Machine Intelligence and the International Journal of Computer Vision. He also serves/served as an Area Chair of top conferences such as ICCV, CVPR, ECCV, NeurIPS and ICLR.
\end{IEEEbiography}




\clearpage

\appendix

\subsection{Setup on Testing Dataset for BFR}\label{sec:evalation_setup_supp}
In the main text, we synthesized a testing dataset named CelebA-Test for the task of blind face restoration (BFR), containing 4,000 HQ images randomly selected from CelebA-HQ~\cite{karras2018progressive}. To synthesize the LQ images of CelebA-Test, we employ the degradation model in Eq.~(17) of the manuscript with the following settings:
\begin{gather*}
    s \in \{4, 8, 16, 20, 24, 28, 32, 36, 40\}, ~ \sigma \in \{1, 5, 10, 15, 20\}, \\
    q \in \{30, 40, 50, 60, 70\}, ~ \theta \in \{0, \frac{1}{4}\pi, \frac{1}{2}\pi, \frac{3}{4}\pi\}, \\
    l_x, l_y \in \{4, 8, 12, 16\},
\end{gather*}
where $l_x$, $l_y$, and $\theta$ control the generation of the blur kernel $\bm{k}$. Specifically, the covariance matrix $\bm{\Sigma}$ of $\bm{k}$ is defined as follows:
\begin{equation}
    \bm{U}=\begin{bmatrix}
        \cos\theta & -\sin\theta \\
        \sin\theta &\cos\theta
    \end{bmatrix},  ~ ~
    \bm{\Lambda}=\begin{bmatrix}
        l_x^2  &0 \\
        0      &l_y^2
    \end{bmatrix},  ~ ~
    \bm{\Sigma} = \bm{U}\bm{\Lambda}\bm{U}^T.
    \label{eq:kernel_generation}
\end{equation}
\begin{figure*}[t]
    \centering
    \includegraphics[width=\linewidth]{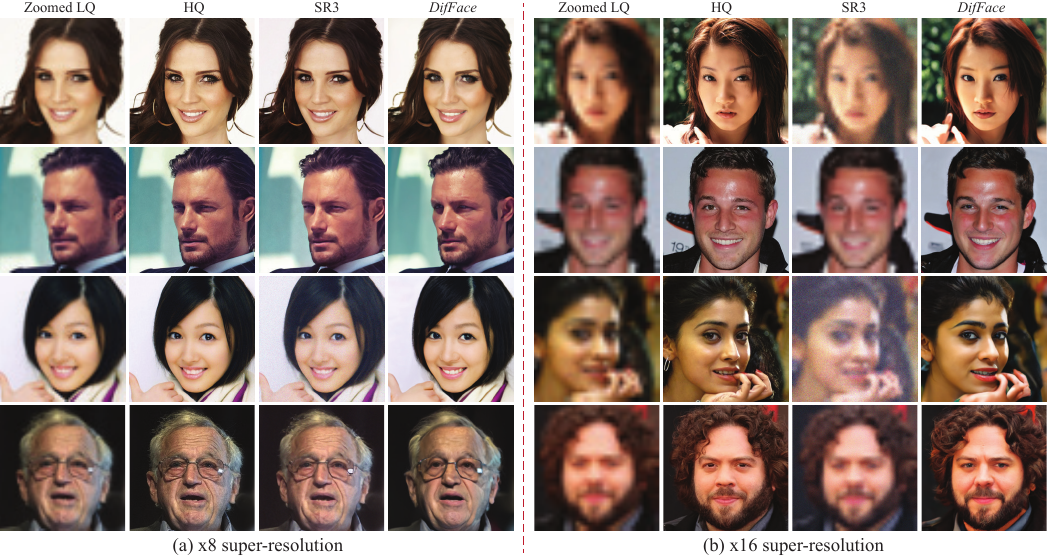}
    \vspace{-5mm}
    \caption{Visual comparisons of \textit{DifFace} and SR3 on 8x (left) and 16x (right) face image super-resolution.}
    \label{fig:bicubic_sr3}
\end{figure*}

\subsection{Network Architecture}\label{sec:network_setup_supp}
\subsubsection{Diffused Estimator for BFR}
We consider two typical restoration backbones for the diffused estimator $f(\cdot;w)$, namely SRCNN~\cite{dong2015image} and SwinIR~\cite{liang2021swinir}. To apply them in our method, we slightly adjust their settings. The LQ and HQ images in our experiments are both with size $512 \times 512$, we add two (or three) PixelUnshuffle~\cite{shi2016real} layers with a downscale factor 2 to reduce the input size to $128 \times 128$ (or $64 \times 64$) for SRCNN (or SwinIR). After each PixelUnshuffle layers except the last one, one convolutional and LeakyReLU layers are followed to fuse the features. Similarly, two (or three) PixelShuffle~\cite{shi2016real} layers are also added to the tail of SRCNN (or SwinIR) to upsample the size back to $512 \times 512$.

As for SRCNN, we adopt nine convolutional layers with kernel size 5 between the PixelUnshuffle and PixelShuffle layers, and each convolutional layer has 64 channels. As for SwinIR, we follow the official settings\footnote{\url{https://github.com/JingyunLiang/SwinIR}} for real-world image super-resolution task with a scale factor 8.

\subsubsection{Diffused Estimator for Inpainting}
The diffused estimator for inpainting is directly borrowed from LaMa~\cite{suvorov2022resolution}, which mainly leverages the technique of fast Fourier convolution (FFC). It takes the low-quality (LQ) image and the image mask as input and outputs the inpainted result. Specifically, the input firstly undergoes a downsampling by a factor of 8 through three sequential FFC layers, each with a stride of 2. Subsequently, the downsampled representation passes through a sequence of 18 FFC-based ResNet~\cite{he2016deep} blocks. Finally, the output is upsampled to attain the desired spatial size via three transposed convolution layers.

\subsection{Experiments on Image Super-Resolution}

While the experiments in this study only focus on blind face restoration and image inpainting, our proposed \textit{DifFace} is a general restoration framework. In this part, we further verify its effectiveness on a more representative restoration task, namely image super-resolution (SR).

\begin{figure*}[t]
    \centering
    \includegraphics[width=\linewidth]{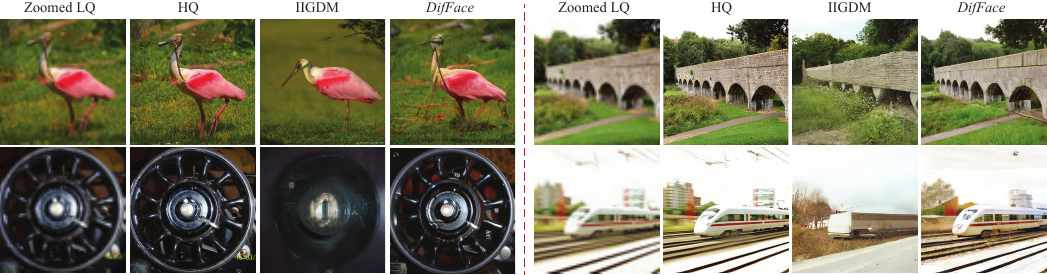}
    \vspace{-5mm}
    \caption{\zsyrevise{Visual comparisons of \textit{DifFace} and IIGDM on 4x (64$\rightarrow$256) natural image super-resolution.}}
    \label{fig:bicsrx4_imagenet}
\end{figure*}
\begin{table*}[t]
    \centering
    \caption{Quantitative results of SR3 and \textit{DifFace} on bicubic face image super-resolution.}
    \label{tab:metirc_bicubic}
    \vspace{-3mm}
    \small
    \begin{tabular}{@{}C{3.4cm}@{}| @{}C{2.4cm}@{}|
                    @{}C{2.0cm}@{} @{}C{2.0cm}@{} @{}C{2.0cm}@{} @{}C{2.0cm}@{}@{}C{2.0cm}@{}}
        \Xhline{0.8pt}
        \multirow{2}*{Methods} & \multirow{2}*{Scale factors} & \multicolumn{5}{c}{Metrics} \\
        \Xcline{3-7}{0.4pt}
                          &                     & PSNR$\uparrow$  & SSIM$\uparrow$  & LPIPS$\downarrow$    & IDS$\downarrow$  & LMD$\downarrow$ \\
        \Xhline{0.4pt}
        SR3               &  \multirow{2}*{x8}  & 22.74           & 0.585           & 0.505                & 25.66      & 1.21       \\
        \textit{DifFace}  &                     & 27.24           & 0.754           & 0.375                & 38.89      & 3.10      \\
        \hline \hline
        SR3               &  \multirow{2}*{x16} & 20.50           & 0.506           & 0.645                & 64.19      & 4.95       \\
        \textit{DifFace}  &                     & 25.03           & 0.712           & 0.412                & 51.36      & 3.57       \\
        \Xhline{0.8pt}
    \end{tabular}
\end{table*}
\begin{table}[t]
    \centering
    \caption{\zsyrevise{Quantitative results of the proposed \textit{DifFace} and IIGDM on x4 (64$\rightarrow256$) natural image super-resolution.}}
    \label{tab:metirc_bicsr_imagenet}
    \vspace{-3mm}
    \small
    \begin{tabular}{@{}C{2.8cm}@{}|@{}C{2.0cm}@{} @{}C{2.0cm}@{} @{}C{2.0cm}@{}}
        \Xhline{0.8pt}
        \multirow{2}*{Methods} & \multicolumn{3}{c}{Metrics} \\
        \Xcline{2-4}{0.4pt}
                          & PSNR$\uparrow$  & SSIM$\uparrow$  & LPIPS$\downarrow$   \\
        \Xhline{0.4pt}
        IIGDM~\cite{song2022pseudoinverse}   & 18.66  & 0.449   & 0.574            \\
        \textit{DifFace}       & 22.05  & 0.548   & 0.338    \\
        \Xhline{0.8pt}
    \end{tabular}
\end{table}

\subsubsection{Face Image Super-resolution}
In fact, our \textit{DifFace} model trained for BFR can be directly applied in face image super-resolution. To achieve this goal, we only upsample the LQ image with nearest interpolation before feeding it into our model. It should be noted that we do not specifically retrain or finetune the diffused estimator for SR.

As for the comparison method, we consider the pioneering work SR3~\cite{saharia2022image} in the field of diffusion-based image super-resolution. It requires retraining the diffusion model from scratch by inserting the LQ image as a condition in each timestep of the diffusion model. Since the code of SR3 is not released, we thus adopt an unofficially re-implemented version\footnote{\url{https://github.com/Janspiry/Image-Super-Resolution-via-Iterative-Refinement}} instead. This model is specifically trained for 8x face image super-resolution from size $64\times 64$ to $512\times 512$. To further evaluate the generalization capability to different degradation, we also test its performance on the task of 16x super-resolution from $32\times 32$ to $512\times 512$. As for the testing dataset, we randomly select 200 images from CelebA-HQ~\cite{karras2018progressive}.

Table~\ref{tab:metirc_bicubic} lists the quantitative results of SR3 and \textit{DifFace} on the task of face image super-resolution, and the corresponding visual comparisons are shown in Fig.~\ref{fig:bicubic_sr3}. \textit{DifFace} achieves better or at least comparable performance on 8x super-resolution, even though SR3 is specifically trained for this bicubic upsampling task. When generalized to 16x super-resolution, \textit{DifFace} outperforms SR3 on both qualitative and qualitative comparisons, indicating the robustness of \textit{DifFace} to unknown degradations. On the other hand, \textit{DifFace} is more efficient than SR3, because SR3 has to pass through the whole reverse process of the diffusion model while \textit{DifFace} starts from the intermediate state (i.e., $\bm{x}_N$) of the reverse Markov chain.

\subsubsection{Natural Image Super-resolution}
In this section, we present experimental results on the natural image super-resolution with a scale factor of 4. Consistent with the settings in BFR, we trained a SwinIR~\cite{liang2021swinir} model using $L_2$ loss as the diffused estimator $f(\cdot;w)$. The unconditional diffusion model\footnote{\url{https://github.com/openai/guided-diffusion}}, introduced by Dhariwal and Nichol~\cite{dhariwal2021diffusion}, was employed as our diffusion prior. It was trained on ImageNet~\cite{deng2009imagenet} at a resolution of $256\times 256$. To thoroughly evaluate the performance of \textit{DifFace}, we conducted comparisons with the recent diffusion-based SR method IIGDM~\cite{song2022pseudoinverse}. For a fair comparison, we used the same unconditional diffusion model for both \textit{DifFace} and IIGDM\footnote{\url{https://github.com/NVlabs/RED-diff}}. The testing dataset comprised 500 images randomly selected from the validation set of ImageNet. We center-cropped these images to a resolution of $256\times 256$ as HQ images and subsequently downsampled them to $64\times 64$ as LQ images.

Table~\ref{tab:metirc_bicsr_imagenet} lists the qualitative comparison results on our testing data set, with the corresponding visual results shown in Fig.~\ref{fig:bicsrx4_imagenet}. \textit{DifFace} significantly surpasses IIGDM in both quantitative and qualitative assessments, indicating its effectiveness in general restoration tasks. Notably, \textit{DifFace} demonstrates a remarkable advantage in preserving fidelity with the LQ image. This fidelity superiority can be attributed to its initialization from the intermediate state of the diffusion model rather than from Gaussian noise.

It is important to note that we evaluated the IIGDM model using an unconditional diffusion model, which does not rely on the annotated label information from ImageNet~\cite{deng2009imagenet}. This setting ensures a fair comparison to \textit{DifFace} and is more applicable to practical scenarios, given the sophistication of annotating a label for each testing image. Under such a setting, the overall performance of IIGDM in our experiments is lower than that reported in the original IIGDM paper, where a conditional diffusion model relying on image labels was employed.

\begin{figure*}[t]
    \centering
    \includegraphics[width=\linewidth]{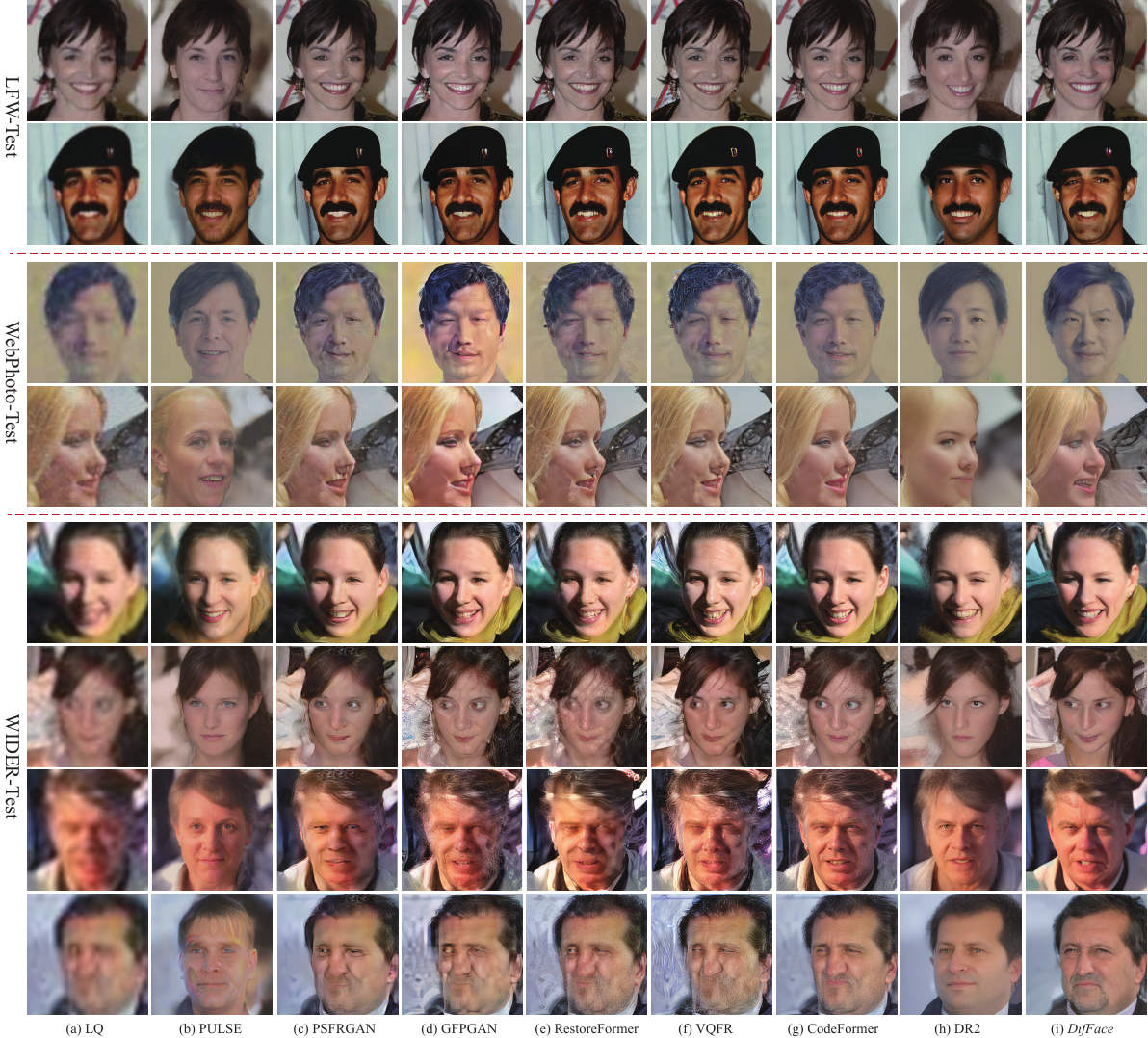}
    \caption{\zsyrevise{Visual comparisons of different methods on three real-world datasets on the task of blind face restoration. Please zoom in for a better view.}}
    \label{fig:real_appedix}
\end{figure*}
\begin{figure*}[t]
    \centering
    \includegraphics[width=\linewidth]{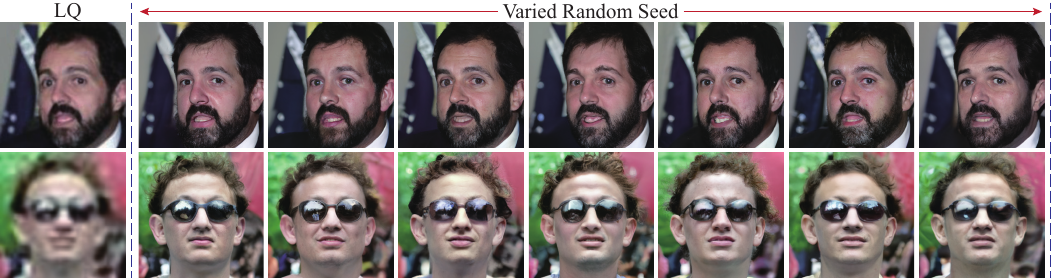}
    \caption{Two restoration examples of \textit{DifFace} on the real-world dataset WIDER-Test
        by setting different random seeds for the diffusion model.}
    \label{fig:random_real_appedix}
\end{figure*}
\begin{figure*}[t]
    \centering
    \includegraphics[width=\linewidth]{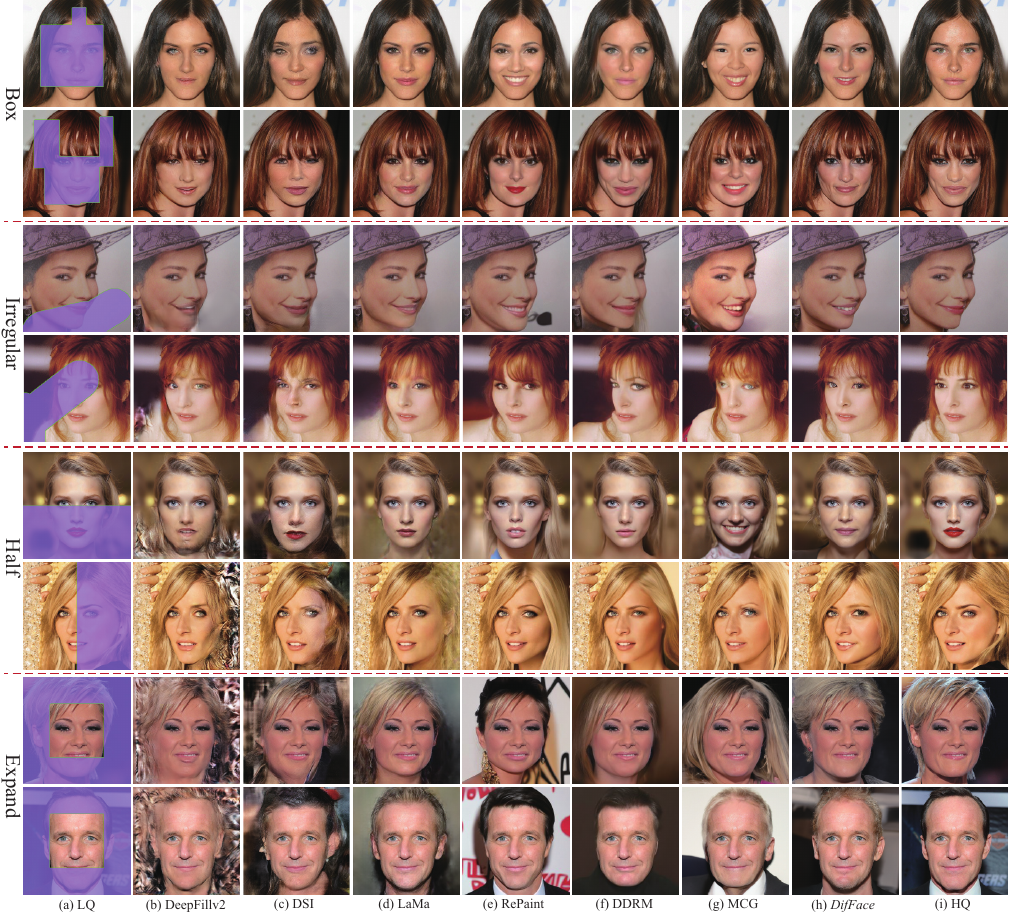}
    \caption{Viusal comparisons of different methods on the task of face inpainting. Please zoom in for a better view.}
    \label{fig:syn8_inpaining_supp}
\end{figure*}

\end{document}